%% file: main.tex
\documentclass[11pt]{article}

\usepackage{amsmath,amsbsy,amsfonts,amssymb,amsthm,dsfont,fullpage,units}
\usepackage{graphicx}
\usepackage{caption}
\usepackage{multirow}
\usepackage{subcaption}
\usepackage{epsfig}
\usepackage{float}
\usepackage[usenames, dvipsnames]{color}
\usepackage{cases}
\usepackage{xcolor}
\usepackage[normalem]{ulem}

\usepackage{algorithmic}
\usepackage[linesnumbered,lined,boxed,commentsnumbered]{algorithm2e}
\usepackage{tikz}
\usetikzlibrary{calc,shapes}
\usepackage{hyperref}
\usepackage{dsfont}
\usepackage{textcomp}

\newtheorem{claim}{Claim}
\newtheorem*{claim*}{Claim}

\newtheorem{lemma}{Lemma}[section]

\newtheorem{theorem}{Theorem}[section]

\theoremstyle{definition}

\newtheorem{definition}{Definition}

\graphicspath{ {diagrams/} }

\include{notations}

\definecolor{WildStrawberry}{RGB}{255,67,164}

\renewcommand{\circ}{\cdot}

\title{Testing Symmetric Markov Chains From a Single Trajectory}

\author {
Constantinos Daskalakis\thanks{Supported by a Microsoft Research Faculty Fellowship, and NSF Award CCF-1551875, CCF-1617730 and CCF-1650733.}\\
EECS, MIT \\
\tt{costis@mit.edu}
\and
Nishanth Dikkala\thanks{Supported by NSF Award CCF-1551875, CCF-1617730 and CCF-1650733.}\\
EECS, MIT \\
\tt{nishanthd@csail.mit.edu}
\and
Nick Gravin\thanks{Supported by NSF Award CCF-1551875, CCF-1617730 and CCF-1650733.} \\
EECS, MIT\\
\tt{ngravin@gmail.com}
}

\begin{document}

\maketitle

\input{abstract}

\thispagestyle{empty}
\addtocounter{page}{-1}
\newpage
\section{Introduction}
\label{sec:intro}
\input{intro}

\section{Preliminaries}
\label{sec:prelim}
\input{prelim}


\section{Distance between Markov Chains}
\label{sec:distance}
\input{distance}

\section{Identity Testing of Symmetric Markov Chains}
\label{sec:symmetric}
\input{cover_time_upper_bound}

\section{A Lower Bound for Identity Testing of Symmetric Markov Chains}
\label{sec:symmetric_lb}
\input{symm_lower_bound}

\section{Open Questions}
\label{sec:open}
\input{openquestions}

\bibliographystyle{alpha}
\bibliography{bib}




%
%
%
%
%

\end{document}

%% file: notations.tex
\newcommand{\be}{\begin{equation}}
\newcommand{\ee}{\end{equation}}
\newcommand{\beq}{\begin{equation*}}
\newcommand{\eeq}{\end{equation*}}

\newcommand{\AutoAdjust}[3]{\mathchoice{ \left #1 #2  \right #3}{#1 #2 #3}{#1 #2 #3}{#1 #2 #3} }
\newcommand{\Xcomment}[1]{{}}

\newcommand{\InBrackets}[1]{\AutoAdjust{[}{#1}{]}}
\newcommand{\Ex}[2][]{\operatorname{\mathbf E}_{#1}\InBrackets{#2}}

\newcommand{\Prx}[2][]{\operatorname{\mathbf{Pr}}_{#1}\InBrackets{#2}}
\newcommand{\Prlong}[2][]{\operatornamewithlimits{\mathbf{Pr}}_{#1}\InBrackets{#2}}

\def\prob{\Prx}

\newcommand{\vect}[1]{\ensuremath{\vec{#1}}}

\newcommand{\eqdef}{\stackrel{\textrm{def}}{=}}
\newcommand{\abs}[1]{\left|#1\right|}

\newcommand{\onenorm}[1]{\left\| {#1} \right\|_1}

\newcommand{\twonorm}[1]{\left\| {#1} \right\|_2}
\newcommand{\infnorm}[1]{\left\| {#1} \right\|_{\infty}}
\newcommand{\iprod}[2]{\langle #1, #2 \rangle}
\newcommand{\scalprod}[2]{\langle #1, #2 \rangle}
\newcommand{\trans}[1]{{#1}^{\top}}

\newcommand{\srprod}[2]{\left[#1,#2\right]_{\text{\textsurd}}}
\newcommand{\srprodt}[2]{\left[#1,#2\right]_{\text{\textsurd}}^{\top}}

\newcommand{\onev}{\vect{\mathds{1}}}
\newcommand{\onevt}{{\trans{\onev}}}
\newcommand{\onevi}[1][i]{\vect{e}_{#1}}
\newcommand{\onevti}[1][i]{{\trans{\onevi[#1]}}}
\newcommand{\specr}[1]{\rho\left(#1\right)}
\newcommand{\eigi}[1][i]{\lambda_{#1}}
\newcommand{\eigvi}[1][i]{\vect{v}_{#1}}
\newcommand{\eigvit}[1][i]{\vect{v}_{#1}^{\top}}
\newcommand{\eigvli}[1][i]{\vect{u}_{#1}}
\newcommand{\eigvlit}[1][i]{\vect{u}_{#1}^{\top}}
\newcommand{\vev}{{\vect{v}}}
\newcommand{\vevt}{{\trans{\vect{v}}}}

\newcommand{\Uniform}[1]{\ensuremath{\mathrm{Unif}\left(#1\right)}}

\newcommand{\hitt}[1]{\mathrm{HitT}_{#1}}
\newcommand{\mixt}[1]{\mathrm{MixT}_{#1}}
\newcommand{\returnt}[1][i]{\mathrm{Return}_{#1}}
\newcommand{\word}[2]{\mathcal{W}_{_{#1}}^{#2}}

\newcommand{\dtvname}{\mathrm{d}_{_\mathrm{TV}}}
\newcommand{\dtv}[2]{\dtvname\left(#1,#2\right)}

\newcommand{\hellingername}{\mathrm{d}_{_\mathrm{Hel}}}
\newcommand{\hellingersqname}{\mathrm{d}_{_\mathrm{Hel}}^{2}}
\newcommand{\hellinger}[2]{\hellingername\left(#1,#2\right)}
\newcommand{\hellingersq}[2]{\hellingersqname\left(#1,#2\right)}

\newcommand{\infmap}[1]{\mathcal{K}_{#1}}
\newcommand{\dist}[2]{\mathrm{Dist}\left(#1,#2\right)}

\newcommand{\Ocomplex}[1]{O\left(#1\right)}
\newcommand{\wO}[1]{\widetilde{O}\left(#1\right)}
\newcommand{\wTheta}[1]{\widetilde{\Theta}\left(#1\right)}
\newcommand{\wOm}[1]{\widetilde{\Omega}\left(#1\right)}

\newcommand{\citep}{\cite}
\newcommand{\citet}{\cite}

\newcommand{\Complex}{\mathbb{C}}
\newcommand\E{\mathbb{E}}
\newcommand\R{\mathbb{R}}
\newcommand\N{\mathbb{N}}

\newcommand\eps{\varepsilon}

\newenvironment{prevproof}[2]{\noindent {\em {Proof of {#1}~\ref{#2}:}}}{$\hfill\qed$\vskip \belowdisplayskip}

%% file: abstract.tex
\label{sec:abstract}
\begin{abstract}
Classical distribution testing assumes access to i.i.d.~samples from the distribution that is being tested. We initiate the study of Markov chain  testing, assuming access to a {\em single trajectory of a Markov Chain.} In particular, we observe a single trajectory $X_0,\ldots,X_t,\ldots$ of an unknown, symmetric, and finite state Markov Chain $\cal M$. We do not control the starting state $X_0$, and we cannot restart the chain. Given our single trajectory, the goal is to test whether  $\cal M$ is identical to a model Markov Chain ${\cal M}'$, or far from it under an appropriate notion of difference.


We propose a measure of difference between two Markov chains, motivated by the early work of Kazakos~\cite{Kazakos78},  which captures the scaling behavior of the total variation distance between trajectories sampled from the Markov chains as the length of these trajectories grows. We provide efficient testers and information-theoretic lower bounds for testing identity of symmetric Markov chains under our proposed measure of difference, which are tight up to logarithmic factors if the hitting times of the model chain ${\cal M}'$ is $\tilde{O}(n)$ in  the size of the state space $n$. 
\end{abstract}

%% file: intro.tex
We formulate theories about the laws that govern physical phenomena by making observations and testing them against our hypotheses. A common scenario is when our observations can be reasonably modeled as i.i.d.~samples from a distribution that we are trying to understand. This is the setting tackled by most classical work in Statistics. Of course, having access to i.i.d.~samples from a distribution is rare and quite commonly an approximation of reality. We typically only have access to approximate samples from a stationary distribution, sampled by a Markov chain whose transition matrix/kernel is unknown to us and which can only be observed for some finite time horizon. In fact, to the best of our knowledge, the underlying Markov chain may not even be rapidly mixing, so there is no guarantee that we will ever see samples that are approximately distributed according to the stationary distribution.

These issues are exacerbated in high-dimensional settings, e.g.~when observing the configurations of a deck of cards where the state space consists of $52!$ permutations, or a weather system, where it may also be completely impractical to work with the high-dimensional stationary distribution itself. 
{Moreover, several different processes may generate the same stationary distribution.} 
For all  these considerations, it may be both more interesting and more practical to understand the ``mechanics'' of the process that generates our observations, namely the transition matrix/kernel of the Markov chain whose evolution we get to observe.

Motivated by these considerations, in this paper we initiate the study of testing identity of Markov chains, and as a first step we focus on the case of finite and symmetric\footnote{
We also get a few observations for general asymmetric case that may be used as a foundation for future studies.} Markov Chains. In 
our setting, we are given access to a {\em single} trajectory $X_0, X_1,\ldots,X_t,\ldots$ of some {\em unknown} symmetric Markov chain ${\cal M}$ over some finite state space $[n]$, and we want 
to test the identity of ${\cal M}$ to some {\em given} symmetric Markov chain ${\cal M}'$ over the same state space. Importantly, we do not get to control the distribution of the starting state 
$X_0$, and we can only observe a single trajectory of ${\cal M}$, i.e. {\em we cannot restart the Markov chain}. What could we hope to achieve?

If there is any difference in the transition matrices of $\cal M$ and $\cal M'$, one would think that we would {\em ultimately} be able to identify it by observing a sufficiently long trajectory. This is certainly true if the transition matrices of the two chains differ at a state that belongs to the essential communicating class (see Definition~\ref{def:comm-class}) of $\cal M$ where $X_0$ lies. However, it is, in general, not always necessary that one be able to observe such a difference. For instance, consider the following example. \\

\noindent \textbf{The Two Communicating Classes Example:} Suppose that $\cal M$ is a chain on states $\{1,2,\ldots,7\}$ whose transition matrix is the random walk matrix on a graph that is the disjoint union of a square on nodes $\{1,\ldots,4\}$ and a triangle on nodes $\{5,6,7\}$, while $\cal M'$'s transition matrix is the random walk matrix on a graph that is the disjoint union of a clique on nodes $\{1,\ldots,4\}$ and a triangle on nodes $\{5,6,7\}$. If our observed trajectory of $\cal M$ lies in the strong connected component defined by states $\{1,\ldots,4\}$ (which forms an essential communicating class), we will easily identify its difference to $\cal M'$. On the other hand, if our observed trajectory of $\cal M$ lies in the essential communicating class defined by states $\{5,6,7\}$, we will not be able to identify that we are not observing a trajectory of $\cal M'$, no matter how long the trajectory is.\\


For some notion of difference, $\dist{\cal M}{\cal M'}$, between Markov chains, we would like to quantify {\em how long} a trajectory $X_0,\ldots,X_{\ell}$ from an {\em unknown} chain, $\cal M$, we need to observe to be able to distinguish, with probability at least $1-\delta$:
\begin{align}
{\cal M}={\cal M}'~~\text{versus}~~\dist{\cal M}{\cal M'}>\epsilon, \label{eq:gof MC testing}
\end{align}
for some given parameters $\delta \in (0,1)$ and $\epsilon>0$. Let us call this problem {\em single-sample goodness-of-fit (or identity) testing for Markov chains}. We will study it taking $\delta=1/3$, with the understanding that this probability can be boosted to any small constant at the cost of a $O(\log (1/\delta))$-multiplicative factor in the length $\ell$ of the observed trajectory.

What notion of difference between Markov chains is the right one to use to study the afore-described goodness-of-fit testing problem? Here are some desiderata for such a notion of difference:
%
%
\begin{enumerate}
\item First, as our simple example above illustrates, under a worst-case starting state $X_0$, we may not be able to identify that ${\cal M} \neq {\cal M}'$ from a single trajectory. So, we would like to identify a notion of difference that takes a value $\dist{\cal M}{\cal M'}= 0$, whenever chains $\cal M$ and $\cal M'$ are indistinguishable from a single trajectory starting at a worst-case starting state.\footnote{The worst-case starting state assumption is a choice also made when defining mixing time. It is also worth noting that in this scenario, since the chains are reducible they will not converge to the stationary distribution and hence the mixing time is infinite.} Obviously, if the chains are irreducible, this constraint is immaterial.

\label{distance property 1}

\item Whenever $\cal M$ and $\cal M'$ {\em are} distinguishable from a single trajectory, whose starting state we do not get to control, i.e.~from any starting state, we would like that our difference measure quantifies {\em how different} the chains are. 
Clearly, our notion of difference could not just be a combinatorial property of the connectivity of the state space of $\cal M$ and $\cal M'$, since the combinatorial structure won't reflect the magnitude of the differences in the chains. 
\label{distance property 2}
\end{enumerate}
One of our main contributions is to identify a meaningful measure of difference between Markov Chains capturing the above properties.

\vspace{-10pt}\paragraph{A Difference Measure Between Markov Chains.}
%
%
%
Total Variation (TV) distance is a standard notion of distance between distributions used in the property/distribution testing literature. One reason for this is that it captures precisely our ability to distinguish
two distributions $p$ and $q$ by observing a single sample from one of them.\footnote{Formally, consider a guessing game where $p$ or $q$ is chosen uniformly at random (or by an adversary), then one sample is generated from the chosen distribution, and we must guess which one it is. The optimal error for this guessing game is precisely $0.5\cdot(1-\dtv{p}{q})$.} 
Similarly, given two product measures $p^{\otimes \ell}$ and $q^{\otimes \ell}$, outputting a vector of $\ell$ i.i.d.~samples drawn from $p$ and $q$ respectively, our ability to distinguish between them from one sample is captured by $\dtv{p^{\otimes \ell}}{q^{\otimes \ell}}$. Unfortunately, it is analytically difficult to relate $\dtv{p^{\otimes \ell}}{q^{\otimes \ell}}$ to $\dtv{p}{q}$ to study how our distinguishing ability improves with $\ell$. For this reason, Hellinger distance $\hellinger{p^{\otimes \ell}}{q^{\otimes \ell}}$ is often the preferred notion of distance in this case.\footnote{Indeed, it enjoys the precise recurrence relation $1-\hellingersq{p^{\otimes \ell}}{q^{\otimes \ell}}=\left[1-\hellingersq{p}{q}\right]^{\ell}$. Moreover, there is a tight relationship between TV and Hellinger distances, see~\eqref{eq:tv hellinger relation}, so one can derive upper and a lower bound on $\dtv{p^{\otimes \ell}}{q^{\otimes \ell}}$ based on $\hellinger{p^{\otimes \ell}}{q^{\otimes \ell}}$. See Section~\ref{sec:prelim}.}.
Generalizing from product measures to Markov Chains, a natural notion of difference between two chains $\cal M$ and $\cal M'$ is the total variation distance, $\dtv{\word{\cal M}{\ell}}{\word{\cal M'}{\ell}}$, 
between $\ell$-step trajectories (a.k.a.~{\em words}) $\word{\cal M}{\ell} \eqdef X_0 X_1 \cdots X_{\ell}$ and $\word{\cal M'}{\ell} \eqdef Y_0 Y_1 \cdots Y_{\ell}$ sampled from the two chains starting at some state $X_0=s_0=Y_0$. 
But due to the analytical difficulties presented by the TV distance for high-dimensional distributions we look towards the Hellinger distance as noted above. The usage of Hellinger square distance for high-dimensional distributions, for instance as was proposed in the early work of Kazakos~\cite{Kazakos78} and the more recent work of Daskalakis and Pan~\cite{DaskalakisP17}, is well known. Hence, we study the Hellinger distance $\hellinger{\word{\cal M}{\ell}}{\word{\cal M'}{\ell}}$ between two trajectories, which satisfies a precise recurrence formula stated as Lemma~\ref{lemma:kazakos lemma} in Section~\ref{sec:distance}. The relation between Hellinger and TV distances allows us to provide upper and lower bounds on the latter in terms of the former.
\vspace{-10pt}\paragraph{A Scale-Free Measure of Difference Between Markov Chains.} Both the distance measures $\dtv{\word{\cal M}{\ell}}{\word{\cal M'}{\ell}}$ and $\hellinger{\word{\cal M}{\ell}}{\word{\cal M'}{\ell}}$ depend on (1): the length $\ell$ of the trajectory and (2): the starting state $s_0$. We would like, instead, a parameter-free and scale-free notion of difference between Markov Chains satisfying the above desiderata. A popular way of tackling such a parameter dependency in Markov Chain literature is to study the inverse dependency of the length $\ell$ of a trajectory required to achieve a certain threshold value for some quantity, e.g.~mixing time is defined as the minimum number of steps $\ell$ needed so that the distribution of the $\ell$-th state of a trajectory starting at any state $s_0$ is no more than $1/4$ away
from the stationary distribution. Similarly, in our case, we propose to analyze the minimum number of steps $\ell$ required so that $\hellinger{\word{\cal M}{\ell}}{\word{\cal M'}{\ell}}$ is at least some constant (we choose $0.5$):\footnote{Note that a trajectory of this length also satisfies $\dtv{\word{\cal M}{\ell}}{\word{\cal M'}{\ell}} \ge 0.25$.} 
\begin{align}
\min_{\ell>0} \quad \ell: \quad\quad \forall s_0\in[n]\quad\quad \hellinger{\word{\cal M}{\ell}}{\word{\cal M'}{\ell}~|~X_0=Y_0=s_0}\ge \delta. \label{eq:smallest_t_for_tv}
\end{align}
\noindent The above definition assumes a worst-case starting state $s_0$ which reflects our desiderata stated above that we do not get to control the starting state and we cannot restart the chain. Moreover, it is the choice made in the definition of mixing time. In Section~\ref{sec:distance} we show a tight relationship between the above definition and an appropriate ``average-case'' version. 
\medskip Clearly, the answer to~\eqref{eq:smallest_t_for_tv} depends on the 
%
{\em scaling behavior}, as $\ell \rightarrow \infty$, of the following quantity:
\begin{align}
\delta(\ell)\eqdef\min_{s_0} \hellinger{\word{\cal M}{\ell}}{\word{\cal M'}{\ell}~|~X_0=Y_0=s_0}. \label{eq:worst word tv}
\end{align}
Interestingly, as we discuss in Section~\ref{sec:prelim}, this scaling behavior is tightly captured by 
the following matrix:
$$\srprod{P}{Q}\eqdef\left[\sqrt{P_{ij}\cdot Q_{ij}}~ \right]_{ij\in[n\times n]},$$
where $P$ and $Q$ are the transition matrices of the two chains, i.e. $P_{ij}$ and $Q_{ij}$ denote the probabilities of transitioning from state $i$ to state $j$ in the two chains. In 
Lemma~\ref{lemma:kazakos lemma}, we state a recursive decomposition that allows us to exactly express the square Hellinger similarity, 
$1-\hellingersq{\word{\cal M}{\ell}}{\word{\cal M'}{\ell}}$ of $\ell$-length words sampled from the two chains in terms of the $\ell$-th power of the above matrix, and the distribution of the 
starting states $X_0$ and $Y_0$ in the two words. 

To identify a word-length independent measure of difference between the two chains based on~\eqref{eq:smallest_t_for_tv}, we employ a spectral approach.
We show that the scaling behavior (w.r.t. $\ell$) of the Hellinger square distance between $\word{\cal M}{\ell}$ and $\word{\cal M'}{\ell}$ is captured by the largest eigenvalue $\lambda_1=\rho(\srprod{P}{Q})$ of matrix $\srprod{P}{Q}$. We show that always $\lambda_1 \le 1$ (Claim~\ref{cl:eigval_less_than_one}), and 
that $\lambda_1=1$ if and only if the two chains have an identical essential communicating class (Claim~\ref{cl:eigval_less_than_one}), in which case we would be unable to identify the difference between the two 
chains from a single trajectory which starts at a state in the essential communicating class which is identical in the two chains (see the two communicating classes example above). These statements hold even for asymmetric chains.
For 
\emph{symmetric} Markov chains,
$\ell$ in~\eqref{eq:smallest_t_for_tv} 
is almost proportional to $\frac{1}{\eps}$ ($\ell=\wTheta{\frac{1}{\eps}}$ up to a $\log n$ factor, see Claim~\ref{cl:symm_spectrum}) where $\eps=1-\rho(\srprod{P}{Q})$.\footnote{
	For non symmetric Markov chains, one can show that the slowest (with respect to the choice of the starting state) that the square Hellinger similarity
	(defined as $1-\hellingersqname$) of the two chains 
	can drop as a function of the length $\ell$ is $\lambda_1^\ell$, up to factors that do not depend on $\ell$; this follows from~\eqref{eq:hellinger_square_algebraic} 
	and~\eqref{eq:largest_eigenvalue}. That is, the slowest that the square Hellinger distance of the two chains can increase is $1-O(\lambda_1^\ell)$. However, the dependency on the
	starting state is more significant than in the symmetric case, and the dependency in the worst-case may be not as smooth as for the symmetric ${\cal M}$ and ${\cal M'}$. (See Figure~\ref{fig:examples} for examples of irregular behavior of certain non-symmetric MC.)
}.
The latter estimation on $\ell$ also holds for the case when initial state in $P$ and $Q$ is chosen uniformly at random.

Given these properties, we propose the use of 
$$\dist{\cal M}{\cal M'} = 1-\rho(\srprod{P}{Q})$$ as a scale-free and meaningful  measure of difference between Markov chains. 
Figure~\ref{fig:examples} illustrates how $\dist{\cal M}{\cal M'}$ behaves for different pairs of Markov chains $\cal M$ and $\cal M'$.

\paragraph{Our Results.} Using our proposed measure of difference between Markov chains we provide algorithms for  goodness-of-fit testing of Markov chains, namely Problem~\eqref{eq:gof MC testing},
%
%
where $\dist{\cal M}{\cal M'}=1-\rho(\srprod{P}{Q})$, where $P$ and $Q$ are the transition matrices of  chains $\cal M$ and $\cal M'$. We study this problem when $\cal M$ and $\cal M'$ are 
both {\em symmetric}, and provide  upper and lower bounds for the minimum length $\ell$ of a trajectory from the unknown chain $\cal M$ that is needed to determine the correct answer with 
probability at least $2/3$. In particular, Theorems~\ref{th:symmetric_ub} and~\ref{thm:symm-lower-bound} combined show that the length of the required trajectory from $\cal M$ to answer 
Problem~\eqref{eq:gof MC testing} is $n/\epsilon$, where $n$ is the size of the state space, up to logarithmic factors and an additive term that does not depend on $\epsilon$ or $\cal M$. Our 
upper bound is established via an information-efficient reduction from single-sample identity testing for Markov chains with $n$ states to the classical problem of identity testing of 
distributions over $O(n^2)$ elements, from i.i.d.~samples. A naive attempt to obtain such a reduction is to look at every $\mixt{\cal M'}$-th step of the trajectory of $\cal M$, where 
$\mixt{\cal M'}$ is the mixing time of chain $\cal M'$, pretending that these transitions are i.i.d.~samples from the distribution $\{\frac{1}{n} P_{ij}\}_{ij \in [n^2]}$. This both incurs an 
unnecessary blow-up of a factor of $\mixt{\cal M'}$ in the required length of the observed trajectory, and is not clear how to analyze rigorously due to the dependencies between samples and the 
fact that the mixing time of $\cal M$ is unknown. We show how to avoid these issues via a more subtle approach, which also exchanges the mutliplicative dependence on the mixing time of $\cal M'$ 
with an additive term that is quasi-linear in the hitting time of $\cal M'$.

\paragraph{Related Work.}
\input{related}

\paragraph{Organization}
We start in Section~\ref{sec:prelim} with a description of the notational conventions we use and provide all necessary formal details for our difference measure in Section~\ref{sec:distance}. 
In Section~\ref{sec:symmetric}, we study the problem of testing identity of symmetric Markov chains and present our tester. We give a sample complexity lower bound for this problem in Section~\ref{sec:symmetric_lb}.

%% file: related.tex
Testing goodness-of-fit for distributions has a long history in Statistics; for some old and more recent references see, e.g.,~\cite{Pearson00,Fisher35,RaoS81,Agresti11}. 
In this literature the emphasis has been on the asymptotic analysis of tests, pinning down their error exponents as the number of samples tends to infinity~\cite{Agresti11,TanAW10}. In 
the last two decades or so, distribution testing has also piqued the interest of theoretical computer scientists~\cite{BatuFFKRW01,Paninski08,LeviRR13,ValiantV14,ChanDVV14, AcharyaDK15,CanonneDGR15,DiakonikolasK16,DaskalakisDSVV13,CanonneRS14,Rubinfeld12,Goldreich11,Canonne15}, where the emphasis, in contrast, has been on minimizing the number of samples required to test hypotheses with a strong control for both type I and type II errors. A few recent works have identified tight upper and lower bounds on the sample complexities of various testing problems~\cite{Paninski08,ValiantV14,AcharyaDK15,DiakonikolasK16}. All of the papers in this vast body of literature assume access to i.i.d. samples from the underlying distribution. 


Some work in Statistics has considered the problem of testing with dependent samples. For instance, \cite{Bartlett51, Moore82, GleserM83, MolinaMPV02} and the references therein study goodness-of-fit testing under Markov dependences. These works study how the classical tests used to perform goodness-of-fit testing with independent samples, perform when there are Markovian dependencies among the samples. \cite{TavareA83} and more recently \cite{BarsottiPR16} study the problem of testing the stationary distribution of Markov chains. \cite{Kazakos78} studies the problem of asymptotically perfect detection (APD) between two Markov chains. All these works focus on the asymptotic regime where the length of the observed trajectories tends to infinity, and study the conditions under which hypothesis testing can be performed successfully or focus on pinning down the error exponents. In the computer science literature, \cite{BatuFRSW13} considered the problem of testing whether a Markov chain is fast mixing or not. They defined a notion of closeness between two random walks starting at different states of the \emph{same} chain, which is different in spirit to the distance notion we define in this work. In particular, their distance is based on the $L_1$ norm of the state distributions attained by starting at two different states $u$ and $v$ and running the chain for $t$ steps. This ignores any differences in trajectory seen along the way and it is apt for their setting as they focus on mixing time which is a trajectory independent property of the chain.


There is a large body of statistical literature on estimating properties and parameters of Markov chains. Mixing time is one such important and well studied parameter
(see, e.g.,~\cite{HsuKS15} and the references therein), as it is useful in designing MCMC algorithms. The question of mixing time estimation is related to but different than the goodness-of-fit 
kernel testing that we perform here.  

%% file: prelim.tex
We list the general notational conventions used in this paper. We denote vectors by small letters 
such as $\vect{v}$ and matrices by capital letters such as $A$,$B$,$P$,$Q$. The $i^{th}$ entry 
of vector $\vect{v}$ is denoted by $v_i$ or $v[i]$ and the $(ij)^{th}$ entry of matrix $A$ ($i^{th}$ row, $j^{th}$ column) is 
denoted by $A_{ij}$ or $A[ij]$; $\vect{e}_i$ denotes the standard basis vector with 
$1$ in its $i^{th}$ coordinate and $0$ elsewhere; $\onev$ denotes the vector of all ones. 
The ``entrywise'' $L_1$ and $L_2$ norms of a matrix  $A$ are respectfully denoted as $\onenorm{A}=\sum_{i,j}|A_{ij}|$ and $\twonorm{A}=\sqrt{\sum_{i,j}A_{ij}^2}$; 
$\specr{A}$ denotes the spectral radius of matrix $A$, i.e., the maximum absolute eigenvalue of $A$. 
The eigenvalues of $A$ are denoted by $\eigi[1], \dots, \eigi, \dots, \eigi[n]$ and 
the respective right eigenvectors by $\eigvi[1],\dots,\eigvi,\dots,\eigvi[n]$ (left 
eigenvectors by $\eigvli[1],\dots,\eigvli[n])$\footnote{If matrix $A$ is not symmetric, we allow $\eigi\in\Complex$ and $\eigvi,\eigvli\in\Complex^n$. Then,
	we will only use $\eigi[1]\in\R$ and $\eigvi[1],\eigvli[1]\in\R^n$.}; for symmetric matrix $A$ we assume that 
$\eigi[1]\ge \dots \ge \eigi \ge \dots \ge \eigi[n]$.\\

\noindent Two popular notions of distance between distributions will be used heavily in this paper. We state their formal definitions below and also specify the relation between them.
\begin{definition}
	\label{def:stat_dist}
	The {\em total variation} and {\em Hellinger} distances between distributions $p,q$ over $[n]$ are defined as : 
	$\dtv{p}{q}\eqdef\frac{1}{2}\sum\limits_{i\in[n]}\abs{p_i-q_i}$;\quad\quad\quad\quad 
	$\hellingersq{p}{q}\eqdef\frac{1}{2}\sum\limits_{i\in[n]}\left(\sqrt{p_i}-\sqrt{q_i}\right)^2= 1-\sum\limits_{i\in[n]}\sqrt{p_i\cdot q_i}$;
	The following relation between these notions of distance is well known (see, e.g., \cite{GibbsS02}):
	\be
	\label{eq:tv hellinger relation}
	\sqrt{2}\cdot\hellinger{p}{q}\ge\dtv{p}{q}\ge\hellingersq{p}{q}. 
	\ee
\end{definition}


\subsection{Markov Chains}
A {\em discrete-time} Markov chain is a stochastic process $\{X_t\}_{t \in \{0,1,\ldots \}}$ over a state space $S$ which satisfies the Markov property: 
the probability of being in state $s$ at time $t+1$ depends only on the state at previous time $t$. In this paper, we
only consider Markov chains with the {\em finite state space} $[n]$. Such Markov chains can be completely specified 
by a $n \times n$ transition matrix (\emph{kernel}) that contains probabilities of transitioning from state $i$ to state $j$ 
in the $i^{th}$ row and $j^{th}$ column, and a description of the distribution of their starting state. The transition matrix has non-negative entries and is a stochastic matrix. We use capital letters $P,Q,M$ to represent Markov 
chains as well as their respective transition matrices. The stationary distribution $\pi$ of a Markov chain $P$ is a distribution over the state space $S$ such that it satisfies $\trans{\pi}\circ P = \trans{\pi}$. 
Another important parameter is the distribution of the starting state $s_0$ which we denote by $\vect{p}$ (for the Markov chain $P$). It may or not may not be the stationary distribution.\\
The state space of a Markov chain can be partitioned into communicating classes which are groups of states reachable from each other with positive probability.
The formal definition of essential communicating classes is as follows.
\begin{definition}[Essential Communicating Classes]
	\label{def:comm-class}
Given a Markov chain $M$ over the state space $[n]$, we define $x \rightarrow y$ if there exists an integer $r>0$ such that $M^r(x,y) > 0$. Similarly, we define equivalence relation 
$x \leftrightarrow y$ iff $x \rightarrow y$ and $y \rightarrow x$. The equivalence classes under relation $\leftrightarrow$ are called communicating classes. Any communicating class $C$ with the property 
that $y$ must be in $C$ for any $x\in C$ and $x \rightarrow y$ is said to be an {\em essential} communicating class\footnote{An essential communicating class can be intuitively thought of as a strong connected component of the underlying directed graph with no outgoing edges.}.
\end{definition}

\subsubsection{Hitting Times and Mixing Times}
Two commonly studied random variables associated with Markov chains which are relevant to this paper are their mixing times and hitting times. 
\begin{definition}[Hitting Time $\hitt{P}$ of a Markov chain $P$]
	Given a Markov chain $P$ over a state space $[n]$, let $s_t$ denote the state at time $t$. The hitting time of $\hitt{P}$ is 
	\begin{align*}
	\hitt{P} = \max_{r,s \in [n]} \{\E\left[\min \{t \ge 0 : s_t = r \text{ given } s_0 = s\}\right] \}
	\end{align*}
\end{definition}

\begin{definition}[Mixing Time $\mixt{P}$ of a Markov chain $P$]
	Given a Markov chain $P$ with a stationary distribution $\pi$ and a starting state distribution $\vect{p}$, 
	\begin{align*}
	\mixt{P} = \max_{\vect{p}}~\min \{t \ge 0 : \onenorm{ P^t \vect{p} - \pi} \le 1/4 \}
	\end{align*}
\end{definition}


%% file: distance.tex
\begin{figure}[H]
	\centering
	\begin{subfigure}[b]{0.45\textwidth}
		\includegraphics[width=0.9\textwidth]{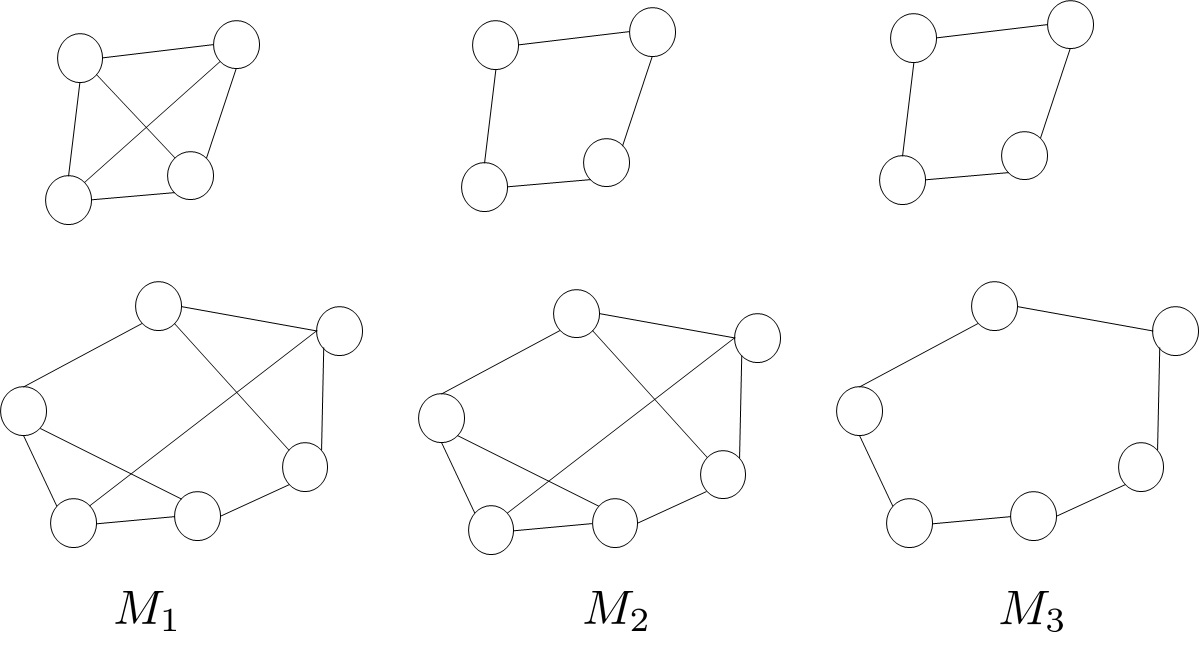}
		\caption{$\dist{M_1}{M_2}=1-\specr{\srprod{M_1}{M_2}}$ is not a metric. $\dist{M_1}{M_2}=\dist{M_2}{M_3}=0,$ but 
			$\dist{M_1}{M_3}> 0$.}
		\vspace{3ex}
		\label{fig:example1}
	\end{subfigure}
	\hspace{5pt}	
	\begin{subfigure}[b]{0.45\textwidth}
		\includegraphics[width=0.9\textwidth]{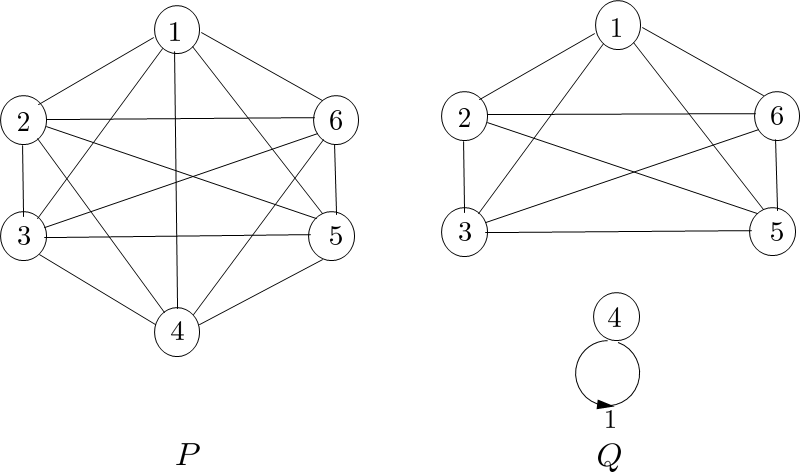}
		\caption{After one step from state $4$, we would know if $w\sim P$, or $w\sim Q$. 
			If $w$ starts from any other state $s_0\neq 4$, it would take many steps.}
		\vspace{3ex}
		\label{fig:example5}
	\end{subfigure}
	
	\begin{subfigure}[b]{0.3\textwidth}
		\includegraphics[width=0.9\textwidth]{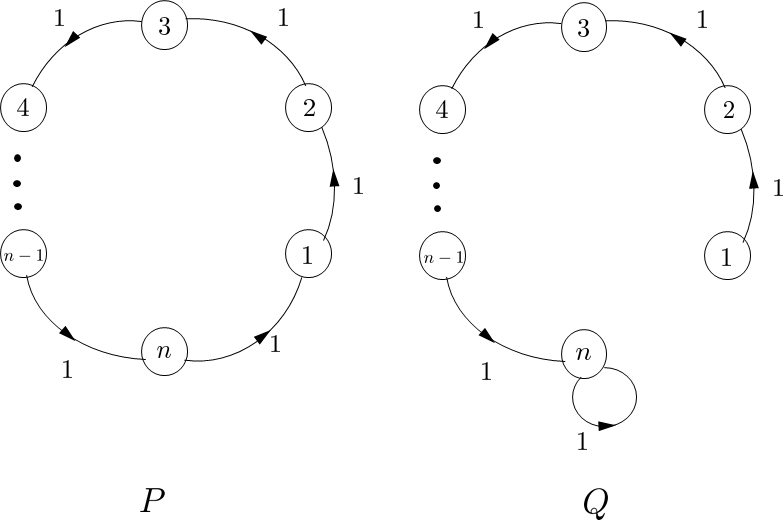}
		\caption{To distinguish $P$ vs. $Q$ walking from a random state we need $\Omega(n)$ steps, but $\dist{P}{Q}=1$.}
		\label{fig:example2}
	\end{subfigure}
	\hspace{3pt}	
	\begin{subfigure}[b]{0.3\textwidth}
		\includegraphics[width=0.9\textwidth]{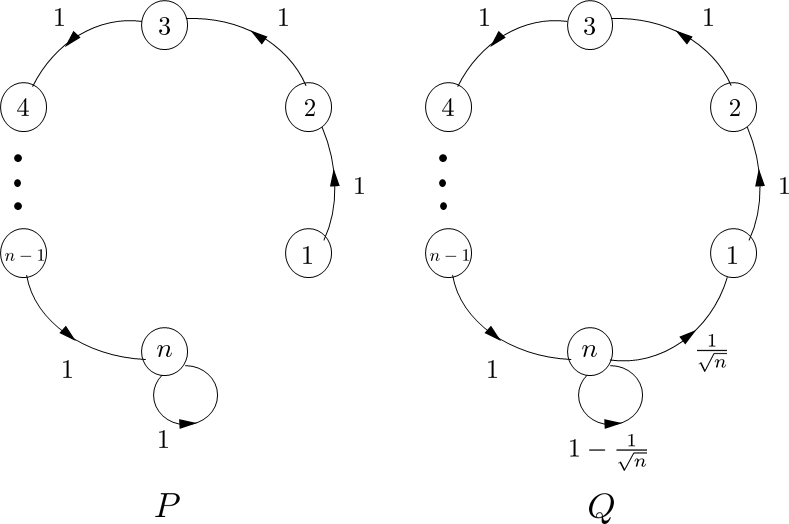}
		\caption{$\dist{P}{Q}=o(1)$, stationary distributions $\vect{q}_0,\vect{p}_0$ 
			are different: $\dtv{\vect{q}_0}{\vect{p}_0}=1-o(1)$.}
		\label{fig:example3}
	\end{subfigure}
	\hspace{3pt}
	\begin{subfigure}[b]{0.3\textwidth}
		\includegraphics[width=0.9\textwidth]{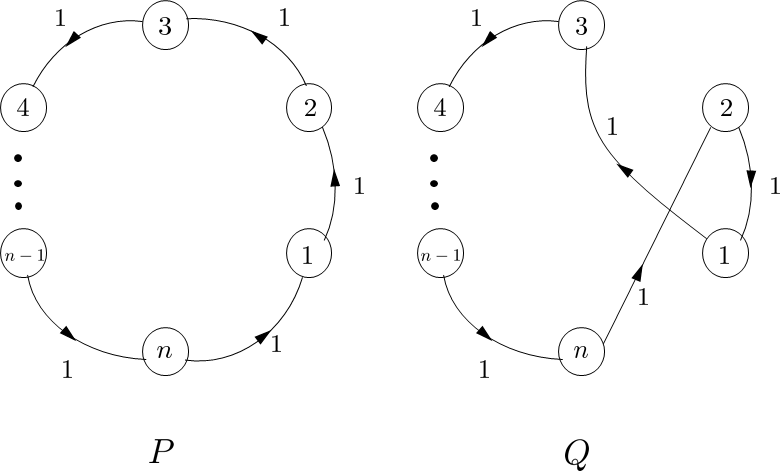}
		\caption{$\dist{P}{Q}=1$. Uniform is stationary	for both $P$ and $Q$. 
			On average $\Omega(n)$ steps to tell $P\neq Q$.}
		\label{fig:example4}
	\end{subfigure}
	
	\vspace{5ex}
	\begin{tabular}{ | l | p{0.8\textwidth} |}
		\cline{2-2}
		
		\multicolumn{1}{c|}{} &  \multicolumn{1}{c|}{Description} \\ \hline
		Example~\ref*{fig:example1} & Two disjoint connected components. \\ \hline
		Example~\ref*{fig:example5} & $Q$ -- clique $K_n$; $P$ -- clique $K_{n-1}$ and disjoint vertex.
		Eigenvalue of $\srprod{P}{Q}$: $\eigi[1]=\sqrt{\frac{n-1}{n}}=1-o(1)$, 
		$\eigi[2]=\sqrt{\frac{1}{n}}$, $\eigi[3]=\cdots=\eigi[n]=0$	\\  \hline												
		Example~\ref*{fig:example2} & $P$ -- oriented cycle, $Q$ -- cycle with one link substituted by a loop.\\ \hline
		Example~\ref*{fig:example3} & $P$ -- oriented cycle with edge $e=(v_1v_2)$ substituted by a loop at $v_1$; 
		$Q$ is almost like $P$, but $e$ has weight $\frac{1}{\sqrt{n}}$, loop at $v_n$ has weight $1-\frac{1}{\sqrt{n}}$. 
		Stationary distributions: $\vect{p}_0=\trans{(1,0,\cdots,0)}$ 
		and $\vect{q}_0=\trans{(\frac{\sqrt{n}}{n+\sqrt{n}-1},\frac{1}{n+\sqrt{n}-1},\dots,\frac{1}{n+\sqrt{n}-1})}$. 
		$\specr{\srprod{P}{Q}}=\sqrt{1-\frac{1}{\sqrt{n}}}$.   \\  \hline
		Example~\ref*{fig:example4} & Two oriented cycles $P\eqdef s_1\to s_2\to\cdots\to s_n\to s_1$ and 
		$Q\eqdef s_1\to s_3\to s_4\cdots\to s_n\to s_2\to s_1$.	\\  \hline											
	\end{tabular}
	
	\caption{Examples.} \label{fig:examples}
\end{figure}

Given two Markov chains $P$ and $Q$, we want to come up with a distance notion which captures how easy it is to distinguish which Markov chain $P$ or $Q$ a word $w=s_0\to s_1\cdots\to s_\ell$ of certain length $\ell$ was generated from (while being agnostic to the distribution of $s_0$). 
This distinguishability is precisely captured by the TV distance $\dtv{\word{P}{\ell}}{\word{Q}{\ell}}$ between {\em word distributions} 
$\word{P}{\ell}$, $\word{Q}{\ell}$ for words of length $\ell$ generated by Markov chains $P$ and $Q$ respectively. It is more convenient in our setting to use, instead of total variation distance, the square of the Hellinger distance $\hellingersq{\word{P}{\ell}}{\word{Q}{\ell}}$
or the closely related Bhattacharya coefficient\footnote{Hellinger distance is tightly related to the Bhattacharya coefficient between two distributions which is defined as
$BC(p,q) = \sum_{i \in [k]} \sqrt{p_i\cdot q_i}$. It captures similarity of two distributions and lies in $[0,1]$.}, which is useful for studying 
divergence of non-stationary and continuous Markov chains as was observed in~\cite{Kazakos78}. \cite{Kazakos78} establishes
nice recurrence relations for the Bhattacharya coefficient of two word distributions, which is captured by the matrix 
$\srprod{P}{Q}\eqdef\left[\sqrt{P_{ij}\cdot Q_{ij}}~ \right]_{i,j\in[n\times n]}$. 
\begin{lemma}[\cite{Kazakos78}] \label{lemma:kazakos lemma}
Suppose $P$ and $Q$ are Markov Chains over states $[n]$, $\vect{p}$ and $\vect{q}$ are probability distributions of the initial state. Let $\word{P}{\ell}$, $\word{Q}{\ell}$ 
be the distributions denoting a length $\ell$ trajectory of Markov Chains $P$ (resp. $Q$) starting at a random node $s_0$ sampled from $\vec{p}$ (resp. $\vec{q}$). Moreover, define 
the vector $\srprod{\vect{p}}{\vect{q}}\eqdef\left[\sqrt{p_s\cdot q_s}\right]_{s\in[n]}$ and the matrix $\srprod{P}{Q}\eqdef\left[\sqrt{P_{ij}\cdot Q_{ij}}~ \right]_{i,j\in[n\times n]}$. Then:
\be
\label{eq:hellinger_square_algebraic}
1-\hellingersq{\word{P}{\ell}}{\word{Q}{\ell}}=\srprodt{\vect{p}}{\vect{q}}\circ \left(\srprod{P}{Q}\right)^{\ell} \circ \onev,
\ee
\end{lemma}
\begin{prevproof}{Lemma}{lemma:kazakos lemma}
	\begin{multline*}
	1-\hellingersq{\word{P}{\ell}}{\word{Q}{\ell}}=\sum_{w=s_0\ldots s_\ell}\sqrt{\Prlong[P]{w}\Prlong[Q]{w}}
	=\trans{\left[\sum_{\substack{w=s_0\ldots s_{\ell}\\s_\ell=s}}\sqrt{\Prlong[P]{w}\Prlong[Q]{w}}\right]}_{s\in[n]}\circ\onev\\
	=\trans{\left[\sum_{r\in[n]}\sqrt{\Prlong[P]{r\to s}\Prlong[Q]{r\to s}}\sum_{\substack{w=s_0\ldots s_{\ell-1}\\s_{\ell-1}=r}}\sqrt{\Prlong[P]{w}\Prlong[Q]{w}}\right]}_{s\in[n]}\circ\onev\\
	=\trans{\left[\sum_{\substack{w=s_0\ldots s_{\ell-1}\\s_{\ell-1}=r}}\sqrt{\Prlong[P]{w}\Prlong[Q]{w}}\right]}_{r\in[n]}\circ
	\begin{bmatrix}
	&\vdots&\\
	\cdots&\sqrt{P_{rs}\cdot Q_{rs}}&\cdots\\
	&\vdots&
	\end{bmatrix}_{r,s\in[n\times n]}
	\circ\onev
	\\
	=\trans{\left[\sum_{\substack{w=s_0\ldots s_{\ell-1}\\s_{\ell-1}=r}}\sqrt{\Prlong[P]{w}\Prlong[Q]{w}}\right]}_{r\in[n]}\circ\srprod{P}{Q}\circ\onev
	=\srprodt{\vect{p}}{\vect{q}}\circ \left(\srprod{P}{Q}\right)^{\ell} \circ \onev,
	\end{multline*}
\end{prevproof}
There are two important parameters which affect the expression given by \cite{Kazakos78}. The first is the distributions of the starting states of the Markov chains ($\vect{p}, \vect{q}$) and 
the second is the length of the word ($l$). We want a notion of distance which is a scale-free non-negative real number. To achieve this, we study next how to eliminate 
the dependencies on the starting state distributions ($\vect{p},\vect{q}$) and the word length ($l$).
\paragraph{Assumption on the starting state.} We study two scenarios for the choice of the starting state: (i) a {\bf worst-case} scenario where
both $P$ and $Q$ begin from the same state $i$ chosen in adversarial manner to make $P$ and $Q$ look as much alike as possible;
(ii) an {\bf average-case} scenario, where the initial distributions $\vect{p}=\vect{q}$ for $P$ and $Q$ either are given to us, or are related to $P$ and $Q$ in 
some natural way\footnote{For example $\vect{p}$ and $\vect{q}$ could be respective stationary distributions of $P$ and $Q$. However, we still assume identical initial distributions for $P$ and $Q$, i.e. $\vect{p}=\vect{q}$, as otherwise there might be a simpler trivial strategy to distinguish $P$ and $Q$ by observing only one initial sample from $\vect{p}$. Example~\ref{fig:example3} illustrates how 
two Markov chains can produce very similar distributions of words $\word{P}{\ell},\word{Q}{\ell}$ starting from any state for some large $\ell$, and yet have vastly 
different stationary distributions.}. 
Given the assumption on the starting state we want to answer the question of what $\ell$ to pick, so 
that $\word{P}{\ell}$ and $\word{Q}{\ell}$ are far apart in squared Hellinger distance (say $\ge 0.5$). 
Formally, we have the following respectively for the worst-case and average-case scenarios listed above:
\begin{align}
\label{eq:forall_states_eigenvalue}
\min_{\ell>0} \quad \ell:& \quad\quad \forall i\in[n] && 0.5 \ge 1-\hellingersq{\word{P}{\ell}}{\word{Q}{\ell}} = \onevti\circ \left(\srprod{P}{Q}\right)^{\ell} \circ \onev .\\
\min_{\ell>0} \quad \ell:  & && 0.5 \ge 1-\hellingersq{\word{P}{\ell}}{\word{Q}{\ell}} = \srprodt{\vect{p}}{\vect{q}}\circ \left(\srprod{P}{Q}\right)^{\ell} \circ \onev \nonumber
\end{align}
Due to the relation between Hellinger and total variation distances, an inequality similar to~\eqref{eq:forall_states_eigenvalue} holds
for $1-\dtv{\word{P}{\ell}}{\word{Q}{\ell}}$ as well but with a different constant on the left.\\

We call the minimal $\ell$ that satisfies $\dtv{\word{P}{\ell}}{\word{Q}{\ell}}\ge\frac{2}{3}$ for all starting states $i\in[n]$ (or for fixed starting distributions $\vect{p}=\vect{q}$)
the {\em minimal distinguishing length}. We note that \eqref{eq:forall_states_eigenvalue} gives us an estimate on $\ell$ up to a constant factor.

Next we argue that when $\ell$ is large, the behavior of the RHS of~\eqref{eq:forall_states_eigenvalue} is governed by {\em the largest eigenvalue} 
$\eigi[1]=\specr{\srprod{P}{Q}}$ of $\srprod{P}{Q}$. 
In particular, by Perron-Frobenius theorem, we have that the largest eigenvalue of $\srprod{P}{Q}$ is non-negative and 
the corresponding left eigenvector $\eigvli[1]: \eigvlit[1]\circ\srprod{P}{Q}=\eigi[1]\cdot\eigvlit[1]$ 
has non-negative coordinates. In particular, if we choose initial distributions $\vect{p} = \vect{q}$ proportional to $\eigvli[1]$, then
\be
\trans{\vect{p}}\circ\left(\srprod{P}{Q}\right)^{\ell}\circ\onev=\eigi[1]^\ell\cdot\scalprod{\vect{p}}{\onev}=\eigi[1]^\ell.
\label{eq:largest_eigenvalue}
\ee

\begin{claim}
It is always true that $\eigi[1]=\specr{\srprod{P}{Q}}\le 1$. Moreover, $\eigi[1]=1$ iff $P$ and $Q$ have an identical essential communicating class.
\label{cl:eigval_less_than_one}
\end{claim} 
\begin{prevproof}{Claim}{cl:eigval_less_than_one}
	Note that $\frac{P+Q}{2}$ is a stochastic matrix that 
	entry-wise dominates matrix $\srprod{P}{Q}$ with non-negative entries. Therefore,
	$
	\eigi[1]\cdot\scalprod{\eigvli[1]}{\onev}=\eigvlit[1]\circ\srprod{P}{Q}\circ\onev\le\eigvlit[1]\circ\left[\frac{P+Q}{2}\right]\circ\onev
	=\eigvlit[1]\circ\onev=\scalprod{\eigvli[1]}{\onev},
	$
	where $\onev$ is vector with all $1$ entries. We get $\eigi[1]\le 1$, since $\scalprod{\eigvli[1]}{\onev}>0$.
	
	For the case of equality, if $P$ and $Q$ have the same essential communicating class $C$, then matrix $\srprod{P}{Q}$ has the same transition 
	probabilities as Markov chains $P$ and $Q$ restricted to the vertices of $C$. We note that $C$ is a stochastic 
	matrix and, therefore, its largest positive eigenvalue is one. Hence, $\specr{\srprod{P}{Q}}\ge\specr{C} = 1.$
	
	If $\specr{\srprod{P}{Q}}=1$, we apply Perron-Frobinius theorem to $\srprod{P}{Q}$ to get  
	that the largest eigenvalue $\eigi[1]=\specr{\srprod{P}{Q}}=1$ has corresponding (left) eigenvector $\eigvli[1]$ with non-negative entries. 
	We observe that $\eigvlit[1]\circ\left(\frac{P+Q}{2}-\srprod{P}{Q}\right)\circ\onev=0$, 
	and all entries of the matrix in this expression are non-negative. This implies that $P_{ij}=Q_{ij}$ for every strictly positive coordinates 
	$i$ of the eigenvector $\eigvli[1]$ and any $j\in[n]$. Since $\eigvlit[1]\circ\srprod{P}{Q}=\eigvlit[1]$, we also have $P_{ij}=Q_{ij}=0$ for any positive 
	coordinate $i$ and zero coordinate $j$ of eigenvector $\eigvli[1]$. Therefore, the set of vertices corresponding to positive coordinates of $\eigvli[1]$ 
	form a component (which might have more than one connected component of $P$ and $Q$) such that $P=Q$ on these vertices.  
\end{prevproof}
We use the quantity $\eps\eqdef 1-\specr{\srprod{P}{Q}}$ as a proxy for the closeness of Markov chains $P$ and $Q$. In particular in \eqref{eq:forall_states_eigenvalue} if $\vect{p} = \vect{q}$ is proportional to $\eigvli[1]$, then 
$\ell\cdot\ln(1-\eps)\le\ln 0.5\implies\ell\ge\frac{\ln 2}{2\eps}$. This shows that in the worst-case we need to observe a trajectory of length at least $\Omega(1/\eps)$  before we can satisfactorily distinguish the two chains.
Note however that, in general, $\ell$ might need to be larger than 
$\Omega(\frac{1}{\eps})$ as is illustrated in Example~\ref{fig:example2}. However, we will see that in the case of symmetric Markov chains we observe a more regular behavior.
In the remainder of this section and the following sections we only consider symmetric Markov chains that avoid such irregular behavior and 
dependency on the starting state.

\paragraph{Word distance between Symmetric Markov Chains.} The stationary distribution for any symmetric Markov chain is the uniform distribution over all states. 
In this case the most natural starting distributions for the average-case part of equation~\eqref{eq:forall_states_eigenvalue} are $\vect{p}=\vect{q}=\frac{1}{n}\onev$.
In this setting of symmetric Markov chains, we can provide sharp bounds on the minimal distinguishing length $\ell$.
\begin{claim}
The necessary and sufficient distinguishing length $\ell$, which allows to distinguish $P$ vs. $Q$ with high probability, 
is $\wTheta{\frac{1}{\eps}}$ (up to a $\log n$ factor), where $\eps=1-\specr{\srprod{P}{Q}}$ under both worst-case and average-case (we assume
$\vect{p}=\vect{q}=\frac{1}{n}\onev$) scenarios for the starting state.
\label{cl:symm_spectrum}
\end{claim}
\begin{prevproof}{Claim}{cl:symm_spectrum}
	We first consider the average-case model for the starting state. Note that $\srprod{P}{Q}$ is a symmetric matrix. 
	Let $\eigvi[1],\dots,\eigvi[n]$ be normalized orthogonal eigenvectors of $\srprod{P}{Q}$, corresponding to 
	real $\eigi[1]\ge\dots\ge\eigi[n]$ eigenvalues. Then for RHS of \eqref{eq:forall_states_eigenvalue} we have
	\be
	\frac{1}{n}\onevt\circ \left(\srprod{P}{Q}\right)^{\ell}\circ\onev=\frac{1}{n}\onevt\circ\left(\sum_{i=1}^n\eigi\cdot\eigvi\circ\eigvit\right)^\ell\circ\onev
	=\sum_{i=1}^n\eigi^\ell\cdot\frac{1}{n}\iprod{\onev}{\eigvi}^2=(*)
	\label{eq:definition_star}
	\ee
	Now we can write an upper and lower bound on $(*)$ in terms of $\eigi[1]^\ell$ (assuming that $\ell$ is even):
	\begin{multline*}
	\frac{\eigi[1]^\ell}{n}=\frac{\eigi[1]^\ell}{n}\twonorm{\eigvi[1]}^2\le\eigi[1]^\ell\cdot\frac{1}{n}\onenorm{\eigvi[1]}^2
	\le(*) \le
	\sum_{i=1}^n\eigi^\ell\cdot\frac{1}{n}\onenorm{\eigvi}^2\le\sum_{i=1}^n\eigi^\ell\cdot\twonorm{\eigvi}^2=\sum_{i=1}^n\eigi^\ell\le n\cdot\eigi[1]^\ell,  
	\end{multline*}
	where in the second inequality we used Perron-Frobenius theorem stating that all coordinates of $\eigvi[1]$ are non negative. 
	Consequently, these bounds imply that $\ell=\Theta\left(\frac{1}{\eps}\right)$ up to a $\log n$ factor, if $\specr{\srprod{P}{Q}}=\eigi[1]=1-\eps$. 
	I.e., $\ell=\wTheta{\frac{1}{\eps}}$.
	
	For the worst-case assumption on the starting state, it is sufficient to show an upper bound $\ell=\Ocomplex{\frac{\log n}{\eps}}$. 
	In this case~\eqref{eq:definition_star} becomes
	\[
	\onevti\circ \left(\srprod{P}{Q}\right)^{\ell}\circ\onev=\sum_{i=1}^n\eigi^\ell\cdot\iprod{\onevi}{\eigvi}\cdot\iprod{\onev}{\eigvi}
	\le\sum_{i=1}^n|\eigi|^\ell\cdot\infnorm{\eigvi}\cdot\onenorm{\eigvi}
	\le\sum_{i=1}^n|\eigi|^\ell\cdot \sqrt{n}\le n^{1.5}\cdot\eigi[1]^\ell,
	\]
	since $\onenorm{\eigvi}\le\sqrt{n}\twonorm{\eigvi}=\sqrt{n}$, and $\infnorm{\eigvi}\le\twonorm{\eigvi}=1$.
\end{prevproof}

We note that, if one could pick the starting state instead of working with average-case or worst-case assumptions of Claim~\ref{cl:symm_spectrum}, then $\ell$ can be much smaller (see Example~\ref{fig:example5}). Claim~\ref{cl:symm_spectrum} gives a strong evidence that $\dist{P}{Q}\eqdef 1-\specr{\srprod{P}{Q}}$ 
is a meaningful and important parameter that captures closeness between $P$ and $Q$. In the following section we will use it as analytical proxy for the distance between 
Markov Chains\footnote{In general this notion of distance should be used with care. One thing about parameter $\dist{P}{Q}=1-\specr{\srprod{P}{Q}}$, is that it is not a metric. In particular, $
\dist{P}{Q}$ violates the triangle inequality ($\dist{M_1}{M_2}=\dist{M_2}{M_3}=0,$ but $\dist{M_1}{M_3}>0$ for some $M_1,M_2,M_3$) as is illustrated by Example~\ref{fig:example1}. 
We note that this problem can only appear for reducible chains, as is shown in Claim~\ref{cl:eigval_less_than_one}. Also it is not always possible to extend the sharp bounds on $\ell$ of 
Claim~\ref{cl:symm_spectrum} from symmetric Markov chains to non-symmetric Markov chains, even if both MC have the uniform distribution as their stationary distribution (see Example~\ref{fig:example4})
}.

%% file: cover_time_upper_bound.tex
After understanding the problem of distinguishing between two given distributions, a next fundamental question is the identity testing problem where the goal is to test whether an unknown distribution $p$ from which we see a stream of samples, coincides with a given hypothesis distribution $q$. 
In this section, we study identity testing of symmetric Markov chains and provide an efficient algorithm (Theorem~\ref{th:symmetric_ub}).  
We begin by giving below a formal statement of the problem:\\

\begin{algorithm}[H]
	\label{def:mc-id-testing}
\KwIn{$\eps > 0$; explicit description of a symmetric Markov chain $Q$; a trajectory $s_1\cdots s_m$ of length $m$ from a symmetric Markov Chain $P$.}
\KwOut{$P=Q$, or $P\neq Q$ if $1-\specr{\srprod{P}{Q}}>\eps$.}
\end{algorithm}

\paragraph{Our approach.}
Identity testing problem with i.i.d. samples, is a well studied problem in the
distribution testing literature. The problem is quite non trivial\footnote{It is studied by a number of works. For instance, see~\cite{BatuFFKRW01,Paninski08,ValiantV14,AcharyaDK15,DiakonikolasK16} (This is not an exhaustive list).} 
and to achieve tight sample complexity one needs to do careful estimations of collisions in observed samples.
Markov chain identity testing appears to be at least as hard as the i.i.d. identity testing problem with the added complication of dependent samples.
To avoid involved analysis of collisions among dependent samples we will instead try to find a black-box reduction of the MC testing problem to identity testing with i.i.d. samples.
A naive attempt at such a reduction proceeds by waiting for a period of mixing time $\mixt{Q}$ of the known Markov Chain $Q$ to get one (potential) i.i.d. sample from the 
stationary distribution of $P$ (in case $P$ has mixed). If the empirical distribution for the number of visits is far from the uniform distribution, we can 
immediately reject $P$ (since if $P=Q$, then $P$ is a symmetric chain and will have the uniform distribution as the stationary) and if it is not, then we would have attained multiple transitions from a sizeable set of nodes and one could hope they contain sufficient signal to distinguish $P$ from $Q$. 
It is non-trivial to extract this signal as the mere fact that we have seen multiple samples from a single node introduces 
dependencies in our samples. That is, two samples from the same node are not independent samples from the transition distribution of that node, if it took only a little time to return to this node. Moreover, this attempt, if it works, will incur a \emph{multiplicative loss} of $\mixt{Q}$ in the sample complexity.\\

\noindent We take a more subtle and involved approach to achieve a successful reduction to the classical setting with i.i.d. samples. Moreover, our reduction yields an algorithm that suffers only an {\em additive loss} of $\wO{\hitt{Q}\cdot\log\left(\hitt{Q}\right)}$ in sample complexity. We reduce the Markov chain problem to the classical identity testing problem with respect to squared Hellinger distance of distributions supported on a domain of size $n^2$. Our result is always as good as the naive approach. Indeed, the hitting time cannot be larger than Mixing time times $n$, but usually it is much smaller (in fact hitting time can be even smaller than mixing time).  
We note that many broad classes of graphs and Markov chains have close to linear hitting times, e.g., expanders, $d$-dimensional grids (which are not expanders). Below we describe how we map samples from a Markov chain to i.i.d. samples from the appropriate distribution.\\

\noindent \textbf{A Mapping From Infinite Words.} Consider a mapping $\infmap{\vect{k}}$ from words of infinite length $w \in W_M^{\infty}$ of an irreducible Markov chain $M$ to $\prod_{i=1}^{n}
[n]^{k_i}$, where $\vect{k}=(k_1,\cdots,k_n)$ is a vector of $n$ non negative integers, as follows. For each infinite word $w=s_1s_2\cdots$ and each state 
$i\in[n]$ we look at the first $k_i$ visits to state $i$ (i.e., at times $t=t_1,\dots,t_{k_i}$ with $s_{t}=i$) and write down the corresponding 
transitions in $w$, i.e., $s_{t+1}$. We note that every state is visited almost surely in $w$, since $M$ is an irreducible finite-state Markov chain. 
Therefore, mapping $\infmap{\vect{k}}$ defines a probability distribution on $\prod_{i=1}^{n}[n]^{k_i}$. Now, crucially, this distribution is independent across 
all different states and/or independent for a particular state $i$ because of the Markov property of Markov chains. Furthermore, a specific transition from a copy of the state $i$ is distributed according to the $i$-th row of the transition matrix $M$.

In Lemma~\ref{lem:large_deviation_bound}, we show that even for a finite length trajectory with length $ m=\widetilde{O}\left(\hitt{Q}\log\left(\hitt{Q}\right)\right. \allowbreak +\left. \frac{n}{\eps}\right)$ \footnote{in this paper, $\widetilde{O}$ always hides poly $\log(n/\eps)$ factors, but not $\hitt{Q}$.} and 
$k_i=O(\Ex{\text{\# visits to } i})=O(\frac{m}{n})$ the mapping $\infmap{\vect{k}}$ is well defined for all 
but a small fraction (in probability) of the words from the distribution $\word{M}{m}$. This effectively allows us, with high probability, to generate a large number of independent samples from the following distribution supported over $[n]\times[n]$: pick uniformly at random a state $i\in[n]$ and then observe a transition from $i$ according to transition probabilities of row $P_i$.
Indeed, to this end, we first simulate $m'=\Theta\left(\frac{m}{\log^2 (n/\eps)}\right)$ i.i.d. samples from $[n]$. These samples describe how many visits an independent sampler would make to state $i \in [n]$. Let 
$\vect{k}$ be the histogram of these $m'$ samples (note that $\max_{i} k_i \le O(m'\log n /n)$ with high probability). 
We apply $\infmap{\vect{k}}$ mapping to our stream of $m$ consecutive samples of Markov chain $P$, which is well defined with high probability. 
Apart from some small probability events ($\max_{i} k_i$ is too large, or $\infmap{\vect{k}}$ is not defined for our choice of $m$) we obtain the desired $m'$ i.i.d. 
samples. 
\begin{lemma} 
	Given an irreducible Markov chain $M$ and the mapping from infinite words $\word{M}{\infty}$ described above, for $m=\wO{\log\left(\hitt{Q}\right)\hitt{Q}}$, then 
	$\Prx{\exists~\text{state}~i~\text{s.t.}~|\{t: i=s_t\in w\}|<\frac{m}{8e\cdot n}}\le\frac{\eps^2}{n}$ where the probability is over the sampling of $\vect{k}$ and word $w$. 
\label{lem:large_deviation_bound}
\end{lemma}
\begin{prevproof}{Lemma}{lem:large_deviation_bound}
	To simplify notations we denote by $\Delta\eqdef 2\hitt{Q}$. By union bound over all states $i$ it is enough to show that $\Prx{|\{t: i=s_t\in w\}|<\frac{m}{8e\cdot n}}\le\frac{\eps^2}{n^2}$ 
	for each fixed state $i$. We can make sure that in the first $\frac{m}{2}$ steps state $i$ is visited at least once with probability 
	at least $1-\frac{\eps^2}{n^3}$. Once we visited state $i$, instead of hitting time for state $i$ we can analyze the {\em return time} $\returnt$ for $i$.
	Note that for symmetric Markov chains $\frac{1}{n}\onev$ (uniform distribution) is a stationary distribution. Therefore, every state appears at average once 
	in every $n$ steps in an infinite word from $\word{Q}{\infty}$. In other terms, the expectation of $\returnt$ for each 
	state $i$ is exactly $n$. By definition of hitting time we have that in $\frac{\Delta}{2}$ steps the probability 
	of reaching a particular state $i$ from any other state $j$ is greater than $1-1/e$ (or any other given constant). It implies that $\Prx{\returnt \ge \frac{\Delta}{2}\cdot C}\le e^{-C}$
	for any $C\in\N$. Indeed, one can show this by induction on parameter $C$. 
	Notice that if the random walk did not return to $i$ after $C-1$ steps it has stopped at some state $j\neq i$. Then for any choice of $j$ by definition of the hitting time 
	the random walk will return to $i$ with probability at least $1/e$ in the next $\frac{\Delta}{2}$ steps. It is not hard to get a similar bound $\Prx{\returnt \ge \Delta\cdot C}\le e^{-C}$ for any $C\ge 1, C\in\R$. To simplify notations we use $X$ to denote the random variable $\returnt$ 
	and $X_1,\dots,X_\ell$ to denote $\ell$ i.i.d. samples of $X$. We have  
	
	\be
	X\ge 0\quad\text{and}\quad\forall C\in\R_{\ge 1},\Prx{X \ge \Delta\cdot C}\le e^{-C} \quad\text{and}\quad \Ex{X}=n.
	\label{eq:ineq_return}
	\ee
	
	We only need to show that $\Prx{X_1+\cdots+X_\ell > m/2} \le \frac{\eps^2}{n^2}$ for $\ell=\frac{m}{8e\cdot n}$.
	To this end, we use a standard technique for large deviations and apply Markov's inequality to the moment generating function of $X_1 + \cdots + X_\ell$,
	
	\be
	\Prx{X_1+\cdots+X_\ell > m/2}=\Prx{e^{\theta\cdot(X_1+\cdots+X_\ell)} > e^{\theta\cdot m/2}}\le\frac{\Ex{e^{\theta\cdot(X_1+\cdots+X_\ell)}}}{e^{\theta\cdot m/2}} \\
	=\frac{\Ex{e^{\theta X}}^\ell}{e^{\theta\cdot m/2}}
	\label{eq:MGF_markov}
	\ee
	We note that given restrictions \eqref{eq:ineq_return} on $X$ maximum of $\Ex{e^{\theta X}}$ for any fixed $\theta>0$ is attained at 
	\[
	X^*\sim
	\begin{cases}
	\Delta\cdot x  & x\in[C_0,\infty) \text{ with probability density function } e^{-x}\\
	0 & \text{with remaining probability }1-e^{-C_0},
	\end{cases}
	\]
	where constant $C_0>1$ is such that $\Ex{X^*}=n$. Indeed, distribution $X^*$ maximizes \eqref{eq:MGF_markov} due to simple variational inequality:
	$\epsilon\cdot e^{\theta\cdot a}+\epsilon\cdot e^{\theta\cdot b}<\epsilon\cdot e^{\theta\cdot (a-c)}+\epsilon\cdot e^{\theta\cdot (b+c)}$ for any $b \ge a \ge c > 0$, and probability mass $\epsilon>0$.
	This inequality allows us to increase $\Ex{e^{\theta\cdot X}}$ and not change $\Ex{X}$ by tweaking density function $f(x)$ of $X$ 
	$f'(a-c)=f(a-c)+\epsilon$, $f'(a)=f(a)-\epsilon$, $f'(b)=f(b)-\epsilon,$ $f'(b+c)=f'(b+c)+\epsilon$, ($f'(x)=f(x)$ for all other $x$), for some $c\le a$. 
	The only time we cannot apply this incremental change is when $X=X^*$.
	
	We have 
	\be
	\label{eq:C0}
	\Ex{X^*}=\Delta(C_0+1)e^{-C_0} = n.
	\ee 
	We set $\theta\eqdef\frac{1}{2\Delta\log\Delta}$ in \eqref{eq:MGF_markov}. Now we are ready to estimate $\Ex{e^{\theta\cdot X}}$.
	To simplify notations we denote $\gamma\eqdef\frac{1}{2\log\Delta}$.
	\begin{multline}
	\Ex{e^{\theta\cdot X}}=1-e^{-C_0}+\int_{C_0}^{\infty}e^{\theta\cdot\Delta\cdot x}\cdot e^{-x}\,\mathrm{d}x=
	1-e^{-C_0}+\int_{C_0}^{\infty}e^{-x\cdot(1-\frac{1}{2\log\Delta})}\,\mathrm{d}x\\
	=1-e^{-C_0}+\frac{e^{-C_0(1-\gamma)}}{1-\gamma}=1+e^{-C_0}\left(\frac{e^{C_0\gamma}}{1-\gamma}-1\right).
	\label{eq:expectation_MGF}
	\end{multline}
	We notice that $\gamma C_0<1$, since from \eqref{eq:C0} we can conclude that $\frac{e^{C_0}}{C_0+1}=\frac{\Delta}{n}\implies C_0<2\log\Delta=1/\gamma$. The last implication can be obtained as follows: for $C_0 > 2.52,$ we have $C_0 - \frac{C_0}{2}\le C_0-\ln(1+C_0)=\ln(\frac{\Delta}{n})$. Now, we can estimate $e^{\gamma C_0}\le 1 + e\cdot\gamma C_0$ in \eqref{eq:expectation_MGF}. Furthermore, since $\gamma<1/2$ we have
	the term $\frac{e^{C_0\gamma}}{1-\gamma}-1$ in \eqref{eq:expectation_MGF} to be at most $2e\gamma(C_0+1)$. With this estimate we continue \eqref{eq:expectation_MGF}
	\be
	\Ex{e^{\theta\cdot X}}\le 1+e^{-C_0}2e\gamma(C_0+1)=1+\frac{e\cdot n}{\Delta\log\Delta}.
	\label{eq:expectation_MGF2}
	\ee
	We apply estimate~\eqref{eq:expectation_MGF2} and formula $\theta=\frac{1}{2\Delta\log\Delta}$ to~\eqref{eq:MGF_markov} to obtain
	\[
	\Prx{X_1+\cdots+X_\ell > m/2}\le \frac{\left(1+\frac{e\cdot n}{\Delta\log\Delta}\right)^\ell}{e^{m/4\Delta\log\Delta}}
	\le\frac{e^{m/8\Delta\log\Delta}}{e^{m/4\Delta\log\Delta}}=e^{\frac{-m}{8\Delta\log\Delta}}<\frac{\eps^2}{n^2},
	\]
	where in the second inequality we used the fact $\left(1+\frac{e\cdot n}{\Delta\log\Delta}\right)^{\frac{\Delta\log\Delta}{e\cdot n}}<e$, and to get the last inequality
	we used $m=\wOm{\Delta\log\Delta}$ (where in $\widetilde{\Omega}$ the hidden dependency is only on  $\log \eps$ and $\log n$).
\end{prevproof}

In the following we present our algorithm for Markov Chain identity testing and provide an upper bound on its sampling complexity.

\begin{algorithm}[H]
\caption{Independent Edges Sampler.}
\label{alg:symmetric_MC}
\SetKwData{Histogram}{Histogram}\SetKwData{Samples}{Samples}\SetKwFunction{IdentityTest}{IdentityTestIID}
\SetKwFunction{Uniform}{Uniform}
\BlankLine
$\vect{k}~\leftarrow$ \Histogram ($\Theta\left(\frac{m}{\log^2 (n/\eps)}\right)$ i.i.d. \Uniform[n] samples) \;
\For{$t\leftarrow 1$ \KwTo $m-1$}{
\lIf{$|\Samples[s_t]|<\vect{k}[s_t]$}{Add $(s_t\to s_{t+1})$ to $\Samples[s_t]$}
}
\eIf{$\exists i, \text{ s.t., } |\Samples[i]|<\vect{k}[i]$}{
\KwRet \textsc{Reject};
}
{
$\Samples\leftarrow\Samples[1]\cup\dots\cup\Samples[n]$\;
\KwRet \IdentityTest($\eps,$ $\{q_{ij}=\frac{1}{n}\cdot Q_{ij}\}_{i,j\in[n]}$, \Samples)\;
}
\end{algorithm}

Algorithm~\ref{alg:symmetric_MC} uses as a black-box the tester of Algorithm 1 of~\cite{DaskalakisKW17}. The following Lemma follows from Theorem 1 of~\cite{DaskalakisKW17}. 
\begin{lemma}
\label{lem:hellinger_idtest}
Given a discrete distribution $q$ supported on $[n]$ and access to i.i.d. samples from a discrete distribution $p$ on the same support, there is a tester which can distinguish whether $p=q$ or $\hellinger{p}{q} \ge \eps$ with probability $\ge 2/3$ using $\Ocomplex{\frac{\sqrt{n}}{\eps^2}}$ samples.
\end{lemma}
As a corollary of Lemma~\ref{lem:hellinger_idtest}, we get a test that can distinguish whether $P=Q$, or $\hellingersq{\frac{1}{n}P}{\frac{1}{n}Q}\ge\eps$ using $m=\Ocomplex{\frac{n}{\eps}}$ i.i.d samples from $\frac{1}{n}P$, which can be viewed as a distribution on a support of size $n^2$. Lemma~\ref{lem:rho_l1_distances} shows that the required distance condition for the i.i.d. sampler is implied by our input guarantee. 

\begin{lemma}
	Consider two symmetric Markov chains $P$ and $Q$ on a finite state space $[n]$. Denote by $\frac{1}{n}P$ the distribution over $n^2$ elements obtained by scaling down every entry of the transition matrix $P$ by a factor $1/n$. We have,
\begin{align}
\frac{1}{2}\sum\limits_{i,j\in[n]}\left(\sqrt{\frac{P_{ij}}{n}}-\sqrt{\frac{Q_{ij}}{n}}\right)^2=
\hellingersq{\frac{1}{n}P}{\frac{1}{n}Q}\ge\eps.
\end{align}
\label{lem:rho_l1_distances}
\end{lemma}
\begin{prevproof}{Lemma}{lem:rho_l1_distances}
	We note that, as $P$ and $Q$ are symmetric matrices, so is $\srprod{P}{Q}$. Thus we have
	\be
	\label{eq:sv_def}
	1-\eps = \specr{\srprod{P}{Q}} = \max_{\twonorm{\vev}=1} \vevt\circ\srprod{P}{Q}\circ\vev.
	\ee
	If we use a particular $\vev=\frac{1}{\sqrt{n}}\onev$ in \eqref{eq:sv_def}, then we get the following inequality.
	\begin{align*}
	1-\eps \ge & \frac{1}{\sqrt{n}}\onevt\circ\srprod{P}{Q}\circ\frac{1}{\sqrt{n}}\onev = \frac{1}{n}\sum\limits_{i,j} \sqrt{P_{ij}\cdot Q_{ij}} =  1-\hellingersq{\frac{1}{n}P}{\frac{1}{n}Q},
	\end{align*}
	which implies $\hellingersq{\frac{1}{n}P}{\frac{1}{n}Q}\ge\eps$.
\end{prevproof}

Finally, the following Theorem~\ref{th:symmetric_ub} gives an upper bound on sampling complexity of Algorithm~\ref{alg:symmetric_MC}.
We note that $\wO{\hitt{Q}}$ samples are necessary for a reduction approach to work. Indeed, if we are to simulate
$n \log n$ i.i.d. samples ($v\to u$, where $v\sim \Uniform[n]$ and $u\sim P_v$), then we shall see all states
$v\in [n]$ at least once with high probability. I.e., the random walk must visit all the states, which would require at the very least 
$\hitt{Q}$ steps in the random walk. On the other hand, our bound of $\wO{\hitt{Q}\cdot\log\left(\hitt{Q}\right)+\frac{n}{\eps}}$ is always better than a naive
bound of $\mixt{Q}\cdot\frac{n}{\eps}$, since $\hitt{Q}< n\cdot\mixt{Q}$ and, in fact, for most of the reasonable MC $\hitt{Q}$ is much less than that.


\begin{theorem}
Given the description of a symmetric Markov chain $Q$ and access to a single trajectory of length $m$ from another symmetric Markov chain $P$,
Algorithm~\ref{alg:symmetric_MC} distinguishes between the cases $P = Q$ versus $1-\specr{\srprod{P}{Q}}>\eps$ with probability at least $2/3$, for  $m=\wO{\hitt{Q}\cdot\log\left(\hitt{Q}\right)+\frac{n}{\eps}}$. 
\label{th:symmetric_ub}
\end{theorem}
\begin{proof}
	In the case $P=Q$, the probability that Algorithm~\ref{alg:symmetric_MC} proceeds to IID tester, i.e., it
	does not reject $P$, because of small number of visits to a state, is at least
	$\Prx{\forall i\in[n]~|\{t: i=s_t\in w\}|>\frac{m}{8e\cdot n}}\cdot\Prx{\forall i: \frac{m}{8e\cdot n}>k_i} \ge 
	\left(1-\frac{\eps^2}{n}\right)\cdot\left(1-\frac{\eps^2}{n}\right)\ge 1-\frac{2\eps^2}{n}$. In the previous estimate, we used Lemma~\ref{lem:large_deviation_bound} 
	to bound $\Prx{\forall i\in[n]~|\{t: s_t\in w, s_t=i\}|>\frac{m}{8e\cdot n}}$, the fact that $\Prx{\frac{m}{8e\cdot n} \le k_i} \le \frac{\eps^2}{n^2}$ (follows from a Chernoff bound), and a union bound.  IID tester then correctly accepts $P=Q$ with probability at least $4/5$.
	Hence, the error probability is at most $1/5+\frac{2\eps^2}{n}<1/3$.
	
	For the case $P\neq Q$, Lemma~\ref{lem:rho_l1_distances} says that if $1-\specr{\srprod{P}{Q}}>\eps$, then distributions passed down to the IID tester 
	$\{p: p_{ij}=\frac{1}{n}P_{ij}\}$ and $\{q: q_{ij}=\frac{1}{n}Q_{ij}\}$ are at least $\eps$ far in Hellinger-squared distance. A black-box application of Lemma \ref{lem:hellinger_idtest} implies a  $\Ocomplex{\frac{n}{\eps}}$ sampling complexity for the IID tester in our case.
	Furthermore,
	random mapping $\infmap{\vect{k}}:\word{P}{\infty}\to p$ (where $\vect{k}$ is a histogram of $m'=\Theta\left(\frac{m}{\log^2 (n/\eps)}\right)$ i.i.d. uniform samples from $[n]$) produces
	$m'$ i.i.d. samples from $p$. Hence, if Algorithm~\ref{alg:symmetric_MC} has sufficient samples from $P$ to define the mapping $\infmap{\vect{k}}$, it would be able to distinguish $p$ and $q$ with probability
	at least $2/3$. On the other hand, if Algorithm~\ref{alg:symmetric_MC} gets finite number of samples which are not sufficient to define the mapping $\infmap{\vect{k}}$, then 
	it correctly rejects $P$ before even running the IID tester.
	
	Thus in both cases the probability of error is at most $1/3$.
\end{proof}

%% file: symm_lower_bound.tex
\label{sec:symm-lower-bound}
In this section we provide an information theoretic lower bound to the identity testing problem on Markov chains defined in Section~\ref{sec:symmetric}. 
\begin{theorem}
\label{thm:symm-lower-bound}
There exists a constant $c > 0$ and an instance of the identity testing problem for symmetric Markov chains (\ref{def:mc-id-testing}) such that any tester on this instance requires
a word of length at least $c\frac{n}{\eps}$ as input to produce the correct output with probability $> 0.99$.
%
\end{theorem}

\begin{prevproof}{Theorem}{thm:symm-lower-bound}
	We use Le Cam's two point method and construct a symmetric Markov chain $Q$ and a class of symmetric Markov chains $\mathcal{P}$ s.t.  
	(i) every $P \in \mathcal{P}$ is at least $\eps$ far from $Q$. That is $1-\specr{\srprod{P}{Q}} \geq \eps$ for any $P \in \mathcal{P}$;
	(ii) there is a constant $c > 0$, s.t. it is impossible to distinguish a word of length $m$ generated by a randomly chosen 
	Markov chain $\bar{P} \sim \mathcal{P}$, from a word of length $m$ produced by $Q$ with probability equal to or greater than $\frac{99}{100}$ for $m\le\frac{cn}{\eps}$.
	To prove (ii) we show that the total variation distance between the $m$-word distributions obtained from the two processes, $Q$ and $\bar{P}$, is small when $m<\frac{cn}{\eps}$ for 
	some constant $c$. We denote distribution of length $m$ words obtained from $Q$ by $\word{Q}{m}$, and from MC $\bar{P}\sim\mathcal{P}$ by $\word{\mathcal{P}}{m}$. 
	We represent symmetric MC as undirected weighted graphs $G=(V,E)$. 
	We allow graph to have multi-edges (this is helpful to provide an intuitive understanding of the lower bound construction and is not 
	essential). We can ultimately remove all multi-edges and give a construction with only simple edges by 
	doubling the number of states.
	
	\begin{description}
		\item[Markov Chain $Q$:] complete double graph on $n$ vertices with uniform weights, i.e., 
		\[
		\forall \quad i\neq j \quad\quad (ij)_1, (ij)_2\in E  \quad\quad  Q_{(ij)_1}=Q_{(ij)_2}=\frac{1}{2(n-1)}.
		\]
		\item[Family $\mathcal{P}$:] for any pair of vertices $i\neq j$ there are two bidirectional edges $(ij)_1$, $(ij)_2$
		with weights randomly (and independently for each pair of $(i,j)$) chosen to be either
		\[
		P_{(ij)_1}, P_{(ij)_2}=\frac{1\pm \sqrt{8\eps}}{2(n-1)},
		\quad\quad\text{ or }\quad\quad
		P_{(ij)_1},P_{(ij)_2}=\frac{1\mp \sqrt{8\eps}}{2(n-1)}.
		\]
		%
		%
	\end{description}
	To make this instance a simple graph with at most one bidirectional edge between any pair of vertices we apply 
	a standard graph theoretic transformation: we make a copy $i'$ for each vertex $i$; for each pair of double edges 
	$e_1=(ij)_1, e_2=(ij)_2$ construct $4$ edges $(ij),(ij'),(i'j),(i'j')$ with weights $w(ij)=w(i'j')=w(e_1)$ and $w(ij')=w(i'j)=w(e_2)$. 
	
	As all Markov chains $Q$ and $P\in\mathcal{P}$ are symmetric with respect to the starting state, we can assume without loss of generality that word $w$ starts from
	the state $i=1$. First, we observe that for the simple graph $2n$-state representation  
	\begin{lemma}
		\label{clm:p-far-from-q}
		Every Markov chain $P \in \mathcal{P}$ is at least $\eps$-far from $Q$.
	\end{lemma}
	\begin{proof}
		For any $P \in \mathcal{P}$, it can be seen that 
		$$\srprod{P}{Q}\circ \onev = \left(\frac{\sqrt{1+\sqrt{8\eps}}+\sqrt{1-\sqrt{8\eps}}}{2}\right)\cdot\onev.$$
		By Perron-Frobenius theorem $\onev$ is the unique eigenvector corresponding to the largest absolute value eigenvalue.  
		Hence, $\specr{\srprod{P}{Q}} = \frac{\sqrt{1+\sqrt{8\eps}}+\sqrt{1-\sqrt{8\eps}}}{2}$ which by Taylor series expansion implies $1-\specr{\srprod{P}{Q}}\ge\eps+\frac{5}{2}\eps^2+o(\eps^2)
		\ge \eps$ for any $\eps<\frac{1}{8}$.
	\end{proof}
	
	We say that a given word $w=s_1\ldots s_m$ from a Markov chain $P$ represented as a multi-edge graph on $n$ states has a $(ij)$ {\em collision}, 
	if any state transition between states $i$ and $j$ (in any direction along any of the edges $(ij)_1,(ij)_2$) occurs more than once in $w$. We now 
	state and prove the following claims about the Markov chain family $\mathcal{P}$.

	\begin{lemma}
		\label{clm:num-collisions}
		Consider a word $w$ of length $m$ drawn from $Q$. The expected number of collisions in $w$ is
		at most $\Ocomplex{\frac{m^2}{n^2}}=\Ocomplex{\frac{1}{\eps^2}}$. 
	\end{lemma}
	\begin{prevproof}{Lemma}{clm:num-collisions}
		Let $I_w(t_1,t_2,(ij))$ indicate the event that in the multi-edge interpretation of the Markov chain $P$, the transition along $(ij)$ edge occurs at times $t_1<t_2$  in $w$. 
		First, we observe that $\Prx{s_{t_1}=s | s_{t_1-1}=x} \leq \frac{1}{n-1}$ and $\Prx{s_{t_2}=s | s_{t_1-1}=x} \leq \frac{1}{n-1}$ for all $x$ and both $s=i$ or $s=j$.
		Thus for any $t_2\ge t_1+2$ by a union bound for all four possible cases of $s_{t_1},s_{t_1+1},s_{t_2},s_{t_2+1}\in\{i,j\}$ we have 
		$$\E\left[I_w(t_1,t_2,(ij)) \right] \leq \frac{4}{(n-1)^4}.$$
		Similarly, for the case $t_2=t_1+1$ we can obtain 
		$$\E\left[I_w(t_1,t_2,(ij)) \right] \leq \frac{2}{(n-1)^3}.$$
		Let $X$ denote the random variable which is equal to the total number of collisions in the word $w$. Then,
		\begin{align*}
		\Ex{X} &= \sum_{t_2 \ge t_1+2}\sum_{i\neq j} \Ex{I_w(t_1,t_2,(ij))} + \sum_{t_1=1}^{m-1}\sum_{i\neq j} \Ex{I_w(t_1,t_1+1,(ij))}\\
		& \le \frac{4}{(n-1)^4}\cdot\frac{m^2}{2}\cdot\frac{n(n-1)}{2}+ \frac{2}{(n-1)^3}\cdot m\cdot\frac{n(n-1)}{2}
		=\Ocomplex{\frac{m^2}{n^2}}
		\end{align*}
	\end{prevproof}
	
	We also consider 3-way collisions which are collisions where there was at least 3 different transition between a pair of states $i$ and $j$ in the word $w$. 
	
	\begin{lemma}
		\label{cl:3way-collisions}
		Consider a word $w$ of length $m$ drawn from $Q$. The probability of $w$ having a $3$-way collision is at most $O(\frac{m^3}{n^4})=o(1).$
	\end{lemma}
	\begin{prevproof}{Lemma}{cl:3way-collisions}
		Similar to the proof of Lemma~\ref{clm:num-collisions} we can give a sharp upper bound on the expected number of $3$-way collisions with the most significant term
		being $\frac{8m^3}{6 (n-1)^6}\cdot\frac{n(n-1)}{2}$, i.e., the expected number of $3$-way collisions is $\Ocomplex{\frac{m^3}{n^4}}$. By Markov inequality we  
		obtain the required bound on the probability of a $3$-way collision.   
	\end{prevproof}
	
	Now consider a typical word $w$ generated by $Q$. 
	As we know from Lemma~\ref{cl:3way-collisions} it has no 3-way collisions and by Markov inequality
	and Lemma~\ref{clm:num-collisions} has at most $O(\frac{1}{\eps^2})$ collisions with high probability. As we show next a typical word $w$ has similar probabilities under $Q$ or 
	$\bar{P}\sim\mathcal{P}$ models.
	
	\begin{lemma}
		\label{clm:no-collision-analysis}
		For $m=O(\frac{n}{\eps})$ at least $\frac{1}{2}$ fraction of words $w=s_1\cdots s_m$ generated by $Q$ satisfy 
		\[
		\frac{1}{2}\cdot\prob[Q]{w}<\prob[\bar{P}\sim \mathcal{P}]{w} < 2\cdot \prob[Q]{w}
		\]
	\end{lemma}
	\begin{prevproof}{Lemma}{clm:no-collision-analysis}
		For each feasible word $w$ in $Q$, i.e., $w$ such that $\prob[Q]{w}>0$
		\[
		\Prlong[Q]{w}=\left(\frac{1}{2(n-1)}\right)^{m-1}
		\quad\quad 
		\Prlong[\bar{P}\sim\mathcal{P}]{w}=\prod_{j>i}\sum_{\bar{P}_{(ij)_1}=\frac{1\pm\sqrt{8\eps}}{2(n-1)}} \bar{P}_{(ij)_1}^{|\{(ij)_1\in w\}|} \cdot\bar{P}_{(ij)_2}^{|\{(ij)_2\in w\}|} 
		\]
		
		First, if $w$ has only one transition along edge $(ij)$, then the corresponding term in $\prob[\bar{P}\sim\mathcal{P}]{w}$ 
		\[
		\sum_{\bar{P}_{(ij)_1}} \bar{P}_{(ij)_1}^{|\{(ij)_1\in w\}|} \cdot\bar{P}_{(ij)_2}^{|\{(ij)_2\in w\}|}=\frac{1}{2}\left(\frac{1+\sqrt{8\eps}}{2(n-1)}+\frac{1-\sqrt{8\eps}}{2(n-1)}\right)=
		\frac{1}{2(n-1)}.
		\]
		From Lemma~\ref{cl:3way-collisions}, we know that probability of a $3$-way collision in $w$ is $o(1)$ under $Q$ model. We observe that for a $2$-way collision $(ij)$
		(a collision which is not a $3$-way collision), the corresponding term in $\prob[\bar{P}\sim\mathcal{P}]{w}$  for the case of transition along two different 
		edges $(ij)_1$ and $(ij)_2$ is
		\[
		\sum_{\bar{P}_{(ij)_1}} \bar{P}_{(ij)_1}^{|\{(ij)_1\in w\}|} \cdot\bar{P}_{(ij)_2}^{|\{(ij)_2\in w\}|}=\frac{1+\sqrt{8\eps}}{2(n-1)}\cdot\frac{1-\sqrt{8\eps}}{2(n-1)}=
		\frac{(1-8\eps)}{4(n-1)^2}.
		\] 
		We call this type of collision {\em type I} collision. For the other case ({\em type II} collisions) of transition along the same edges the respective probability is 
		$\frac{(1+8\eps)}{4(n-1)^2}$. By Lemma~\ref{clm:num-collisions} and by Markov inequality the total number of collisions is $O(\frac{1}{\eps^2})$ with probability $\frac{3}{4}$. 
		We can also make sure that out of these collisions number of type I and type II collisions is roughly the same. More precisely, the difference between numbers of 
		type I and type II collisions is at most $O(\frac{1}{\eps})$ with probability of at least $\frac{3}{4}$. Indeed, the choice of edge collision type in $w$ is uniform between type I and type II, and is
		independent across all collision edges. Now, for small enough $m$ we can make sure that at least $\frac{1}{2}$ fraction of words $w$ has number of collisions at most 
		$\frac{c_1}{\eps^2}$ and the difference between number of type I and II collisions is at most $\frac{c_2}{\eps}$, for some small constants $c_1,c_2>0$. Thus the corresponding density 
		functions can be related as follows.
		\[
		2>\left(1+8\eps\right)^{\frac{c_2}{\eps}}>\frac{\prob[\bar{P}\sim\mathcal{P}]{w}}{\prob[Q]{w}}>\left(1-64\eps^2\right)^{\frac{c_1}{2\eps^2}}\cdot \left(1-8\eps\right)^{\frac{c_2}{\eps}}>1/2
		\]
		%
	\end{prevproof}

	Lemma~\ref{clm:no-collision-analysis} shows that $\dtv{\word{Q}{m}}{\word{\mathcal{P}}{m}} \leq \frac{3}{4}$, which implies that
	no algorithm can successfully distinguish $Q$ from the family $\mathcal{P}$ with probability greater than $\frac{3}{4}$ for some
	$m=\Omega(\frac{n}{\eps})$.
\end{prevproof}

%% file: openquestions.tex
In this paper, we proposed a new framework for studying property testing questions on Markov chains. 
There seem to be multiple avenues for future research and abundant number of open problems arising from this framework. 
We first list some questions which may be of interest here.
\begin{enumerate}
\item What is the optimal sample complexity for identity testing on symmetric Markov chains? In this paper, we show an upper bound of $\wO{\hitt{Q}\cdot\log\left(\hitt{Q}\right)+\frac{n}{\eps}}$ samples (Theorem \ref{th:symmetric_ub}). We conjecture that $\Theta\left( \frac{n}{\eps}\right)$ (same as our lower bound) is the right sample complexity for this problem and an explicit dependence on the hitting time of chain $Q$ may not be necessary. It is implicitly captured to an extent by the guarantee we get from the parameter $\eps$.

\item As there is a natural operation of taking a convex combination of Markov chains, it is natural to ask how our spectral definition of distance 
$1-\specr{\srprod{P}{Q}}$ between two symmetric chains changes if we substitute either $P$ or $Q$ with a convex combination of $P$ and $Q$. How does the distance now relate to the original value?
\item How is the difference parameter $\eps=1-\specr{\srprod{P}{Q}}$ between two Markov chains $P$ and $Q$ related to the difference between Markov chains
$P^k$ and $Q^k$,i.e., states in Markov chains $P$ and $Q$ being observed only at intervals of size $k$? 
\item Given $\eps_2 \ge \eps_1$, and access to words from each of two chains, can we distinguish whether the two chains are $\le \eps_1$-close or $\ge \eps_2$-far? This problem, known as two-sample testing in literature, is another interesting direction using our framework.
\end{enumerate}

%

%% file: main.bbl
\newcommand{\etalchar}[1]{$^{#1}$}
\begin{thebibliography}{MMPV02}

\bibitem[ADK15]{AcharyaDK15}
Jayadev Acharya, Constantinos Daskalakis, and Gautam Kamath.
\newblock Optimal testing for properties of distributions.
\newblock In {\em Advances in Neural Information Processing Systems 28: Annual
  Conference on Neural Information Processing Systems 2015, December 7-12,
  2015, Montreal, Quebec, Canada}, pages 3591--3599, 2015.

\bibitem[Agr12]{Agresti11}
Alan Agresti.
\newblock {\em Categorical Data Analysis}.
\newblock Wiley, 2012.

\bibitem[Bar51]{Bartlett51}
Maurice~S Bartlett.
\newblock The frequency goodness of fit test for probability chains.
\newblock In {\em Mathematical Proceedings of the Cambridge Philosophical
  Society}, volume~47, pages 86--95. Cambridge Univ Press, 1951.

\bibitem[BFF{\etalchar{+}}01]{BatuFFKRW01}
Tugkan Batu, Eldar Fischer, Lance Fortnow, Ravi Kumar, Ronitt Rubinfeld, and
  Patrick White.
\newblock Testing random variables for independence and identity.
\newblock In {\em Proceedings of the 42nd Annual IEEE Symposium on Foundations
  of Computer Science}, FOCS '01, pages 442--451, Washington, DC, USA, 2001.
  IEEE Computer Society.

\bibitem[BFR{\etalchar{+}}13]{BatuFRSW13}
Tu{\u{g}}kan Batu, Lance Fortnow, Ronitt Rubinfeld, Warren~D Smith, and Patrick
  White.
\newblock Testing closeness of discrete distributions.
\newblock {\em Journal of the ACM (JACM)}, 60(1):4, 2013.

\bibitem[BPR16]{BarsottiPR16}
Flavia Barsotti, Anne Philippe, and Paul Rochet.
\newblock Hypothesis testing for {M}arkovian models with random time
  observations.
\newblock {\em Journal of Statistical Planning and Inference}, 173:87--98,
  2016.

\bibitem[Can15]{Canonne15}
Cl{\'e}ment~L Canonne.
\newblock A survey on distribution testing: Your data is big. but is it blue?
\newblock In {\em Electronic Colloquium on Computational Complexity (ECCC)},
  volume~22, pages 1--9, 2015.

\bibitem[CDGR16]{CanonneDGR15}
Cl{\'e}ment~L. Canonne, Ilias Diakonikolas, Themis Gouleakis, and Ronitt
  Rubinfeld.
\newblock Testing shape restrictions of discrete distributions.
\newblock In {\em Proceedings of the 33rd Symposium on Theoretical Aspects of
  Computer Science}, STACS '16, pages 25:1--25:14, 2016.

\bibitem[CDVV14]{ChanDVV14}
Siu-On Chan, Ilias Diakonikolas, Gregory Valiant, and Paul Valiant.
\newblock Optimal algorithms for testing closeness of discrete distributions.
\newblock In {\em Proceedings of the Twenty-Fifth Annual ACM-SIAM Symposium on
  Discrete Algorithms}, pages 1193--1203. Society for Industrial and Applied
  Mathematics, 2014.

\bibitem[CRS14]{CanonneRS14}
Cl{\'{e}}ment~L. Canonne, Dana Ron, and Rocco~A. Servedio.
\newblock Testing equivalence between distributions using conditional samples.
\newblock In {\em Proceedings of the Twenty-Fifth Annual {ACM-SIAM} Symposium
  on Discrete Algorithms, {SODA} 2014, Portland, Oregon, USA, January 5-7,
  2014}, pages 1174--1192, 2014.

\bibitem[DDS{\etalchar{+}}13]{DaskalakisDSVV13}
Constantinos Daskalakis, Ilias Diakonikolas, Rocco~A. Servedio, Gregory
  Valiant, and Paul Valiant.
\newblock Testing \emph{k}-modal distributions: Optimal algorithms via
  reductions.
\newblock In {\em Proceedings of the Twenty-Fourth Annual {ACM-SIAM} Symposium
  on Discrete Algorithms, {SODA} 2013, New Orleans, Louisiana, USA, January
  6-8, 2013}, pages 1833--1852, 2013.

\bibitem[DK16]{DiakonikolasK16}
Ilias Diakonikolas and Daniel~M. Kane.
\newblock A new approach for testing properties of discrete distributions.
\newblock In {\em Proceedings of the 57th Annual IEEE Symposium on Foundations
  of Computer Science}, FOCS '16, pages 685--694, Washington, DC, USA, 2016.
  IEEE Computer Society.

\bibitem[DKW17]{DaskalakisKW17}
Constantinos Daskalakis, Gautam Kamath, and John Wright.
\newblock Which distribution distances are sublinearly testable?
\newblock {\em arXiv preprint arXiv:1708.00002}, 2017.

\bibitem[DP17]{DaskalakisP17}
Constantinos Daskalakis and Qinxuan Pan.
\newblock Square {H}ellinger subadditivity for {B}ayesian networks and its
  applications to identity testing.
\newblock In {\em Proceedings of the 30th Conference on Learning Theory
  ({COLT})}, 2017.

\bibitem[Fis35]{Fisher35}
Ronald~A. Fisher.
\newblock {\em The Design of Experiments}.
\newblock Macmillan, 1935.

\bibitem[GM{\etalchar{+}}83]{GleserM83}
Leon~J Gleser, David~S Moore, et~al.
\newblock The effect of dependence on chi-squared and empiric distribution
  tests of fit.
\newblock {\em The Annals of Statistics}, 11(4):1100--1108, 1983.

\bibitem[Gol11]{Goldreich11}
Oded Goldreich.
\newblock A brief introduction to property testing., 2011.

\bibitem[GS02]{GibbsS02}
Alison~L Gibbs and Francis~Edward Su.
\newblock On choosing and bounding probability metrics.
\newblock {\em International statistical review}, 70(3):419--435, 2002.

\bibitem[HKS15]{HsuKS15}
Daniel~J Hsu, Aryeh Kontorovich, and Csaba Szepesv{\'a}ri.
\newblock Mixing time estimation in reversible {M}arkov chains from a single
  sample path.
\newblock In {\em Advances in neural information processing systems}, pages
  1459--1467, 2015.

\bibitem[Kaz78]{Kazakos78}
Dimitri Kazakos.
\newblock The {B}hattacharyya distance and detection between {M}arkov chains.
\newblock {\em {IEEE} Trans. Information Theory}, 24(6):747--754, 1978.

\bibitem[LRR13]{LeviRR13}
Reut Levi, Dana Ron, and Ronitt Rubinfeld.
\newblock Testing properties of collections of distributions.
\newblock {\em Theory of Computing}, 9(8):295--347, 2013.

\bibitem[M{\etalchar{+}}82]{Moore82}
David~S Moore et~al.
\newblock The effect of dependence on chi squared tests of fit.
\newblock {\em The Annals of Statistics}, 10(4):1163--1171, 1982.

\bibitem[MMPV02]{MolinaMPV02}
I~Molina, D~Morales, L~Pardo, and I~Vajda.
\newblock On size increase for goodness of fit tests when observations are
  positively dependent.
\newblock {\em Statistics \& Risk Modeling}, 20(1-4):399--414, 2002.

\bibitem[Pan08]{Paninski08}
Liam Paninski.
\newblock A coincidence-based test for uniformity given very sparsely sampled
  discrete data.
\newblock {\em IEEE Transactions on Information Theory}, 54(10):4750--4755,
  2008.

\bibitem[Pea00]{Pearson00}
Karl Pearson.
\newblock On the criterion that a given system of deviations from the probable
  in the case of a correlated system of variables is such that it can be
  reasonably supposed to have arisen from random sampling.
\newblock {\em Philosophical Magazine Series 5}, 50(302):157--175, 1900.

\bibitem[RS81]{RaoS81}
Jon~N.K. Rao and Alastair~J. Scott.
\newblock The analysis of categorical data from complex sample surveys:
  Chi-squared tests for goodness of fit and independence in two-way tables.
\newblock {\em Journal of the Americal Statistical Association},
  76(374):221--230, 1981.

\bibitem[Rub12]{Rubinfeld12}
Ronitt Rubinfeld.
\newblock Taming big probability distributions.
\newblock {\em XRDS: Crossroads, The ACM Magazine for Students}, 19(1):24--28,
  2012.

\bibitem[TA83]{TavareA83}
Simon Tavare and Patricia M.~E. Altham.
\newblock Serial dependence of observations leading to contingency tables, and
  corrections to chi-squared statistics.
\newblock {\em Biometrika}, 70(1):139--144, 1983.

\bibitem[TAW10]{TanAW10}
Vincent~Y.F. Tan, Animashree Anandkumar, and Alan~S. Willsky.
\newblock Error exponents for composite hypothesis testing of {M}arkov forest
  distributions.
\newblock In {\em Proceedings of the 2010 IEEE International Symposium on
  Information Theory}, ISIT '10, pages 1613--1617, Washington, DC, USA, 2010.
  IEEE Computer Society.

\bibitem[VV14]{ValiantV14}
Gregory Valiant and Paul Valiant.
\newblock An automatic inequality prover and instance optimal identity testing.
\newblock In {\em Proceedings of the 55th Annual IEEE Symposium on Foundations
  of Computer Science}, FOCS '14, pages 51--60, Washington, DC, USA, 2014. IEEE
  Computer Society.

\end{thebibliography}
